# Data-driven models for production forecasting and decision supporting in petroleum reservoirs

Mateus A. Fernandes[1,2,3,*], Michael M. Furlanetti[1,2], Eduardo Gildin[3], Márcio A. Sampaio[2]

**Abstract**

Forecasting production reliably and anticipating changes in the behavior of rock-fluid systems are the main challenges in petroleum reservoir engineering. This project proposes to deal with this problem through a data-driven approach and using machine learning methods. The objective is to develop a methodology to forecast production parameters based on simple data as produced and injected volumes and, eventually, gauges located in wells, without depending on information from geological models, fluid properties or even details of well completions and flow systems. Initially, we performed relevance analyses of the production and injection variables, as well as conditioning the data to suit the problem. A fundamental step when working with time series is the definition of observation windows, both for use as inputs and outputs of forecasting algorithms and to form training sets. As reservoir conditions change over time, concept drift is a priority concern and require special attention to those observation windows and the periodicity of retraining, which are also objects of study. For the production forecasts, we study supervised learning methods, such as those based on regressions (with and without regularization) and Artificial Neural Networks, including their variants such as LSTM (Long Short-Term Memory) and CNN (Convolutional Neural Network), to define the most suitable for our application in terms of performance and complexity. In a first step, we evaluate the methodology using synthetic data generated from the UNISIM III compositional simulation model. Next, we applied it to cases of real plays in the Brazilian pre-salt, with promising results, but which require refinement of techniques in the next stages of the project. The expected result is the design of a reliable predictor for reproducing reservoir dynamics, with rapid response, capability of dealing with practical difficulties such as restrictions in wells and processing units, and that can be used in actions to support reservoir management, including the anticipation of deleterious behaviors, optimization of production and injection parameters and the analysis of the effects of probabilistic events, aiming to maximize oil recovery.

**Keywords**: Machine learning. Petroleum reservoirs. Forecasting. Decision support.

[1] Dept. of Reservoir and Production Engineering, Petrobras
[2] Dept. of Mining and Petroleum Engineering, University of São Paulo
[3] Dept. of Petroleum Engineering, Texas A&M University
* Contacting author: matfernan@tamu.edu

## 1.1 Introduction

Obtaining reliable production forecasts is one of the main challenges of petroleum reservoir engineering. Accurately estimating future production is fundamental for both short- and long-term activities, including operational planning, dimensioning equipment and facilities, scheduling offloads, evaluating economic viability of future projects, and estimating reserves. Additionally, anticipating depreciatory reservoir behaviors – that could result in production loss – is essential to enable mitigating and preventive actions within the scope of reservoir management.

These tasks are, however, very complex and their results will always be mere approximations of reality based on simplified representations of the rock-fluid system. This is a consequence of the impossibility of accurately knowing the *in-situ* properties of reservoirs.

### 1.1.1 Traditional forecasting methods in reservoir engineering

The traditional methods to perform forecasting tasks for petroleum reservoirs can be divided into different categories, with applicability defined by their complexities, dependencies on previous information, use of computational resources, and capability to represent phenomena and provide detailed results. Decline analysis, materials balance, and numerical simulation represent the main categories; these methods are widely used and explored in a vast literature, as (AHMED, 2005), (DAKE, 1998), and (FANCHI, 2006). Below, we present an introductory description of these methods.

- The decline curves analysis (ARPS, 1945) can be considered the seminal analytical method for production forecast. It consists of simply fitting a decline equation to a reference historical curve and can be used for individual wells or for the entire reservoirs. The generalized decline model can be represented by the hyperbolic curve (Equation 1), but the harmonic and exponential variants may be appropriate in many cases (THAKUR, 2017).

$$q_t = \frac{q_i}{(1+b \cdot d_i \cdot t)^{1/b}} \quad (1)$$

To fit the model and calculate the production rate at a given time qt, we need to set the initial rate $q_i$, the decline rate $d_i$, and the factor $b$ ($0<b<1$) to minimize the difference in relation to a reference curve. This reference curve may come from the well or reservoir itself (if there is a representative production history) or they may come from cases identified as analogous. In specific situations, e.g. unconventional shale reservoirs, it may be more assertive to use hybrid declines to prevent overestimation of cumulative production, as the reservoir tends to an exponential decline ($b$=0 in Equation 1) after some time (BELYADI et al., 2019).

Due to its simplicity, this method is preferred in mature fields, where the trends in production are consolidated and well-known. However, significant changes in production conditions, such as reaching pressures below saturation and new injections or breakthroughs will not be correctly represented.

- The material balance analysis is a more complete analytical method. It dates back to the seminal work by Schilthuis in the first half of the 20$^{th}$ century, who proposed an equation to relate produced and injected volumes at a given static pressure (DAKE, 1998). This equation is based on measured volumes at surface conditions (corrected to reservoir pressure considering fluids expansion factors and gas release from the oil), rock compressibility, and eventually the influences of a gas cap and water influx from an aquifer.

  Expressing the material balance equation as a straight line, as introduced by Havlena and Odeh and detailed by (DAKE, 1998) we can match production and pressure data to obtain an analytical model of the reservoir. This is useful for forecasting while the drive mechanism is unaltered and will show a deviation from the straight line if the drive mechanism changes.

  The proper application of the material balance requires production history data, a dataset of static pressures, PVT (pressure, volume, temperature) properties of the fluids, and petrophysical information as pore volume and compressibility. The method, however, considers the reservoir as a single block of spatially invariant properties, what will limit its applicability to reservoirs with small to medium volumes and with no relevant heterogeneities.

- The numerical simulation represents a big step up in terms of complexity. It's based on a discretized and more detailed representation of the reservoir, where a grid that may contain millions of cells is populated with rock and fluid properties, estimated for every cell. That requires data from core analysis, well logging, formation tests, seismic acquisition, etc., and extensive work on these to generate a 3D model using fundamentals of disciplines that include geology, reservoir engineering, and statistics.

  Once the model is obtained, porous flow, mass conservation and state equation laws are solved numerically at each simulated time step to reproduce the behavior of the reservoir. It is important to emphasize that the accuracy in the reproduction of this behavior is intrinsically related to the quality of the data that generate the model. Thus, a more complex method or a more refined grid will not necessarily be better for a forecast if they are not supported by representative data.

  When we have a production period completed, the model can be refined by history matching, reducing the uncertainties of the parameters estimated in the construction of the model (FANCHI, 2006). However, this is a type of highly complex inverse problem, where virtually infinite combinations of parameters may result in an adjusted history, but which may not be suitable to faithfully reproduce future behavior.

  In the case of large volume and highly complex reservoirs, the modeling, adjustment, and simulation tasks can be extremely costly in terms of data acquisition, demand for computational resources, and time both for specialists' work and for simulation runs. These simulation runs can last from a few hours to a few days, even when using computer clusters with high processing capabilities. That is a frequent and relevant problem in the oil and gas industry, as reported by Bogachev (2018) and Rios et al. (2020). A common approach to deal with that is scaling the model, making the grid cells larger and fewer, to reduce simulation times. However, that comes with the drawback of making results less accurate.

### 1.1.2 Data-driven methods in reservoir engineering

With those traditional forecast methods in mind, we can think of data-driven and machine learning methods to fill some gaps among them – not to replace them. That is exemplified in Table 1, that shows a comparison of the characteristics of the different groups.

**Table 1 – Capabilities and limitations of forecasting methods used in petroleum engineering.**

| Requirements & Features | Decline Analysis | Material Balance | Numerical Simulation | Data-driven Methods |
|---|---|---|---|---|
| Need for measured and production data | Medium | Medium | High | High |
| Need for rock and fluid data | Low | Medium | High | Low |
| Computational resources demand | Low | Low | High | Medium |
| Accuracy on forecasting by well | High | Low | High | High |
| Capability of anticipating new events | N/A | Low | High | Medium |
| Detailing properties by cell | N/A | N/A | High | N/A |

Data-driven models are becoming increasingly present in the industry (PANDEY, 2020), specially involved in tasks where detailed information on fluid configurations and pressures at specific points in a reservoir are not necessary, or when a large database of production and measurements in wells is available, but petrophysical and/or PVT data are not.

The use of such tools to assist reservoir management is quite relevant, especially thinking about supporting routine decisions such as valve openings and closings, management of production and injection rates, and choosing the best well prioritization strategies. Data-based analyzes can then improve reservoir management in the short and medium-term if the tools developed are adequate, as shown in the workflow in Figure 1.

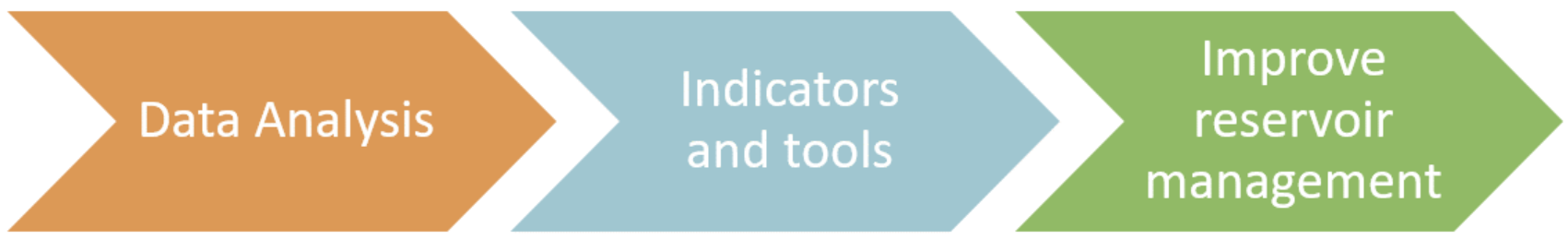

**Figure 1 – General workflow of a data analysis process.**

The main objective of this work is to forecast the production of oil/gas reservoirs using data-driven machine learning models trained only with their production history, without using any information from geology or rock-fluid properties. As secondary objectives, process the production variables and study them to select the most relevant ones and the ideal format for use, in addition to evaluating some machine learning techniques and choosing the most suitable for the main objective, according to their complexity and performance when reproducing the behavior of a reservoir.

### 1.1.3 Literature Review

In the literature we find a variety of data-driven and machine learning methods applied to reservoir engineering problems, as they become increasingly widespread and reliable. In the next paragraphs we bring some examples of prominent work in this area.

(LIU et al., 2020), in order to achieve a fast and accurate production forecast, proposed a learning paradigm with a LSTM (Long Short-Term Memory) neural network based on EEMD (ensemble empirical mode decomposition). The authors establish a machine learning approach based on genetic algorithms that provide a method for predicting oil production in two oil fields in China, with results superior to those presented by Support Vector Machines and Artificial Neural Networks, with lower errors and higher coefficient determination.

(WANG et al., 2019) developed a deep sequential neural network to estimate the accumulated oil production from Bakken reservoirs. They give details of the pre-processing steps, including normalization and data partitioning, and present the deep learning model that after training was able to successfully determine the cumulative production for the studied reservoir. Additionally, Sobol’s analysis of the results showed which interactions are most relevant within the dataset, a valuable information for engineers.

(KUBOTA and REINERT, 2019) proposed two different data-driven approaches – an adaptive linear model and a RNN (recurrent neural network) – to predict future rates for an oilfield subjected to joint water and steam injection. In the case of a field with a long production history and more than 2000 wells drilled, the models dismissed any reservoir information and used as inputs only total injection rate and the number of production wells in operation, presenting as output the total oil rate. Forecasts based on different time frames (both for inputs and outputs) were evaluated, and the results were always satisfactory.

Production forecast for oilfields under waterflood is studied by Deng and Pan (2021) and by Haghshenas et al. (2021). Both rely on data-driven modeling to create proxies that can be used replacing a simulation in specific tasks. The former uses them to forecast water breakthrough, and the latter to optimize injection rates aiming to increase the recovery factor.

Guo and Reynolds (2019) present a data-driven model that takes into consideration fundamentals of the physics of the porous media flow and the fluids transport. By using streamlines connecting nodes that are positioned with the aid of an algorithm in a simplified version of the reservoir, the authors manage to obtain optimal control parameters to maximize the NPV (net present value), instead of using a complete numerical simulation.

Optimization of parameters in a waterflood project, assisted by a model based on data using machine learning techniques, is studied by Jia et al. (2020). The authors obtained additional knowledge of the injection efficiency in each reservoir layer, what was useful to help defining the best schedule for future injection in a reservoir with high geological complexity.

Temirchev et al. (2020) propose a proxy model based on an autoencoder to replace the full reservoir model. This proxy consists of a deep neural network trained with multiple realizations of the simulation model, capable of retaining useful information of the multiphase flow dynamics in the reservoir while requiring only a fraction of the processing time spent by the complete 3D model simulation.

## 1.2 Workflow and methodology

In this work, we use as databases for training and evaluating forecast methods only the produced and injected volumes, in the form of time series with daily discretization and segmented by well. We use supervised learning methods, from which we aim to obtain the mapping between input and output variables by extracting knowledge from the available data. When output variables are continuous, the problem is called regression (PANDEY et al. 2020).

Finding this map, or algorithm, capable of representing the relationships between input and output variables is the human task in the process. The subsequent task is to teach how to go this way many times so that the error in the output variables is minimal, in what we call training step in the machine learning framework. After that, a reserved part of the dataset is used to evaluate the performance of the proposed map, in the step called validation, that prevents us from having a model that is underfitted or overfitted to the training set.

The solutions we propose are based in multilinear and neural network regression models, that were chosen for their ability to deal with time series subject to influences from several concomitant time series, and without periodicity, seasonality, or stationarity. We implemented the algorithms in *Python*, assisted by libraries specialized in data processing, machine learning (*scikit learn*), time series analysis (*pandas*) and data visualization (*matplotlib*). In the next sections we give the details of the workflow, that is represented in Figure 2.

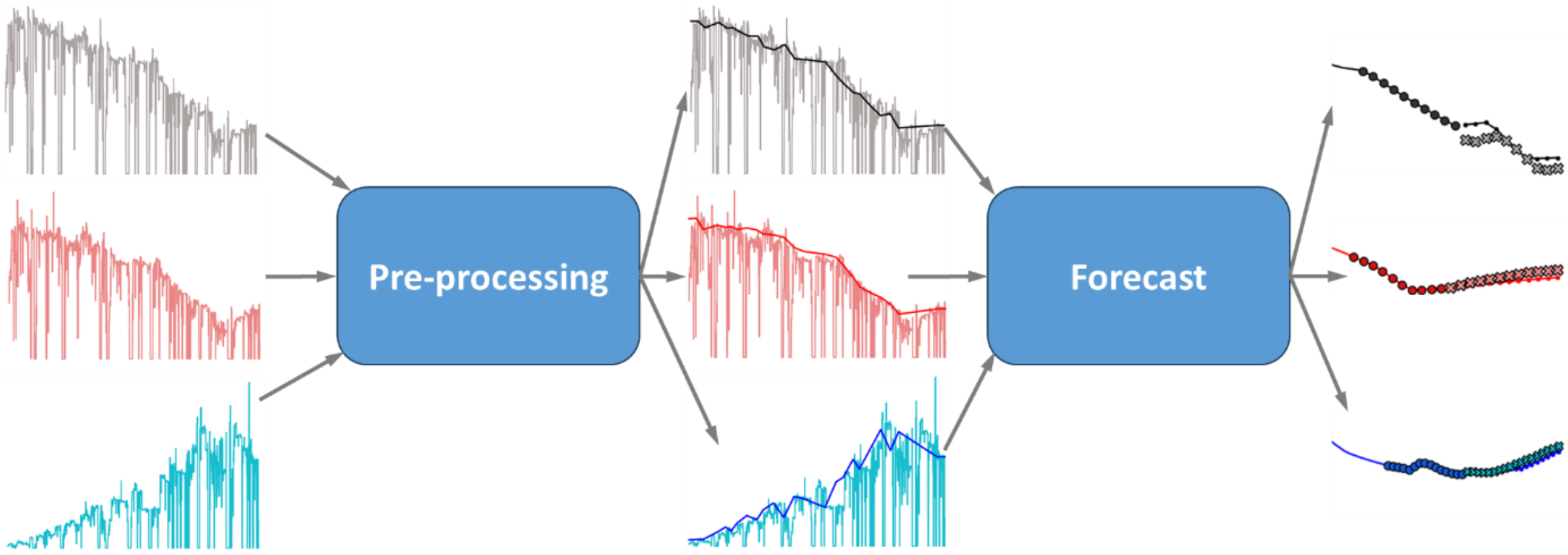

**Figure 2 – Workflow of a data-driven forecast method.**

### 1.2.1 Data conditioning

The initial steps in data conditioning consist of exporting the production curves (the way we obtain these curves is particular to the application, as we will discuss later in our case studies), loading into the *Python* environment, formatting, and inspecting. To avoid inconsistencies in training, the ramp-up period – while new wells are still starting to operate – has been excluded, so all the time windows have the same number of wells in operation. Eliminating this bias is important because we don't consider information *a priori* for new wells, such as their location, or geological and completion data. The daily data generated in the simulations can then be converted into mean values for longer periods, allowing forecast with weekly or monthly discretization, for example. This is important to provide the flexibility to make the database more consistent with the dynamics of the reservoir under study and to make it more suitable to work with the time windows that we require for forecasts of greater practical usage.

The next step is to reshape our dataset as required by our forecast model. This process is illustrated in Figure 3, considering as example the forecast for an entire field. The blue boxes represent the input variables for an observation of the previous *i* time steps and the gray boxes represent the estimated outputs for k future time steps. Time t represents the moment of the first estimate. The production variables $Q_o$, $Q_g$ and $Q_w$ represent oil, gas and water flow rates, respectively; $Q_{gi}$ is the injected gas rate and $Q_{wi}$ the injected water rate. A fundamental part of the task is the choice of the "look back" and "look forward" windows to be used, defining how many time steps in the past will be used as inputs for the forecast and how many time steps in the future will be estimated in each iteration. The evaluation of different configurations to define the ones with best performance is part of the scope of this work.

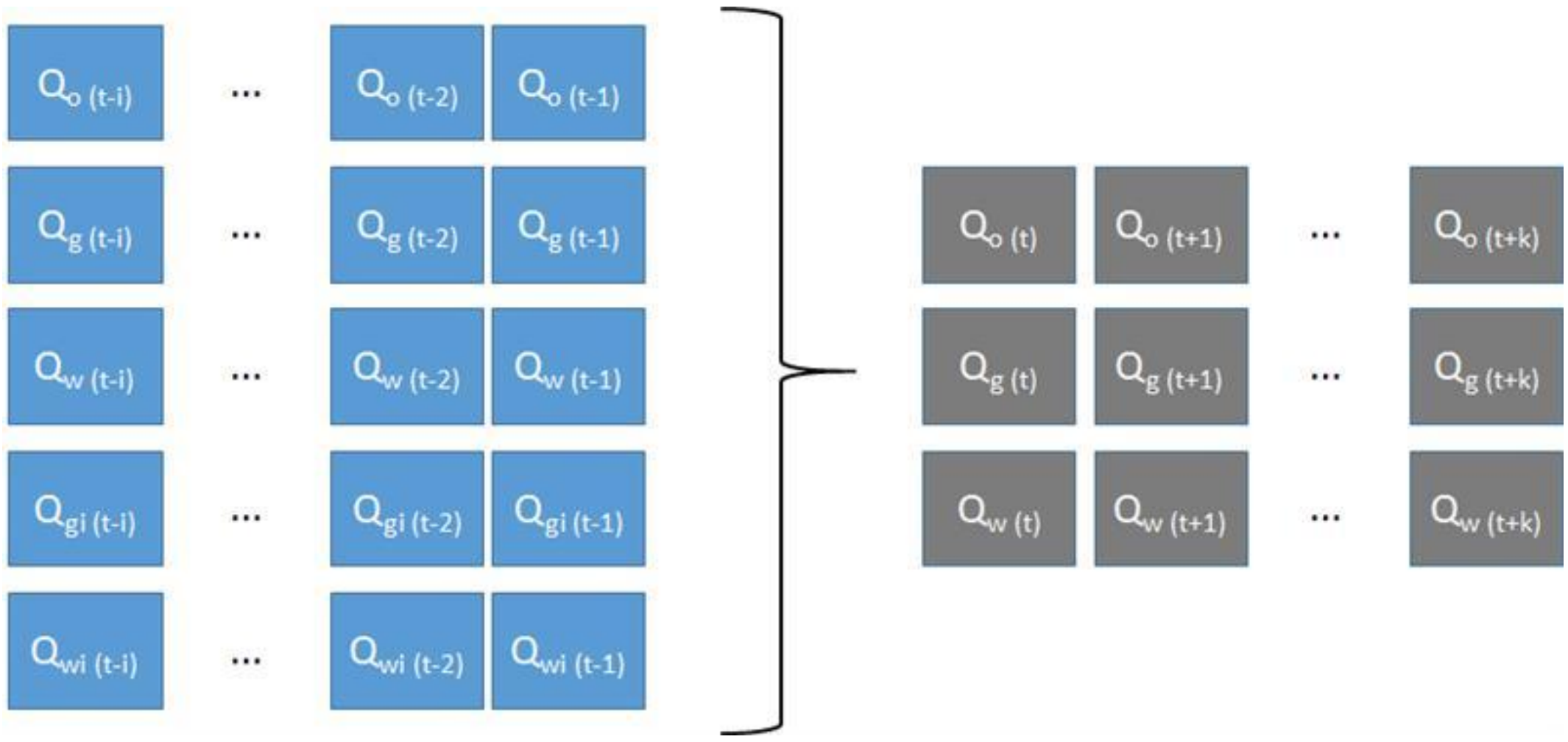

**Figure 3 – Reshaping the dataset to use it in the forecasts; blue boxes represent the inputs and the gray boxes the outputs for the regressors.**

For a forecast segmented per well, each of the variables ($Q_o$, $Q_g$ and $Q_w$ for producers and $Q_{wi}$ or $Q_{gi}$ for injectors) are considered to be in the range [$t$-1, $t$-$i$]. The outputs will be the variables associated with each producing well in the interval [$t$, $t$+$k$]. Thus, the number of inputs and outputs for a problem with the number of producer wells $N_{w_prod}$ and the number of injector wells $N_{w_inj}$ will be given by:

$$N_{inputs} = i * (3 * N_{w_prod} + N_{w_inj}) \quad (2)$$

$$N_{outputs} = k * (3 * N_{w_prod}) \quad (3)$$

We show in Figure 4 an example of the process of reshaping the dataset, as previously described. In the first part, we have the oil rates for three wells; in the second part the new configuration, intended to be used for training and testing. The colors help to show the rearrangement for the first time step.

The dataset obtained is divided into training, validation, and testing subsets. It is important to note that we are dealing with time series, so this division cannot be random; data prior to our reference date will be considered as training and after this date will be used for validation and testing (NIELSEN, 2020). After that, we perform Gaussian normalization, calculating mean and standard deviation only for the training set and using the same values to frame the other subsets, so we do not carry any influence of future information to the learning process.

| Oil rates | | | |
|---|---|---|---|
| Timestep | P1 Qo | P2 Qo | P3 Qo |
| 1 | 10 | 130 | 250 |
| 2 | 20 | 140 | 260 |
| 3 | 30 | 150 | 270 |
| 4 | 40 | 160 | 280 |
| 5 | 50 | 170 | 290 |
| 6 | 60 | 180 | 300 |
| 7 | 70 | 190 | 310 |
| 8 | 80 | 200 | 320 |
| 9 | 90 | 210 | 330 |
| 10 | 100 | 220 | 340 |

| | Training Inputs (x_train) | | | | | | | | | Training Outputs (y_train) | | | | | | | | |
|---|---|---|---|---|---|---|---|---|---|---|---|---|---|---|---|---|---|---|
| Timestep | P1 Qo t-3 | P1 Qo t-2 | P1 Qo t-1 | P2 Qo t-3 | P2 Qo t-2 | P2 Qo t-1 | P3 Qo t-3 | P3 Qo t-2 | P3 Qo t-1 | P1 Qo t | P1 Qo t+1 | P1 Qo t+2 | P2 Qo t | P2 Qo t+1 | P2 Qo t+2 | P3 Qo t | P3 Qo t+1 | P3 Qo t+2 |
| Training only | - | - | - | - | - | - | - | - | - | - | - | - | - | - | - | - | - | - |
| | - | - | - | - | - | - | - | - | - | - | - | - | - | - | - | - | - | - |
| | - | - | - | - | - | - | - | - | - | - | - | - | - | - | - | - | - | - |
| 4 | 10 | 20 | 30 | 130 | 140 | 150 | 250 | 260 | 270 | 40 | 50 | 60 | 160 | 170 | 180 | 280 | 290 | 300 |
| 5 | 20 | 30 | 40 | 140 | 150 | 160 | 260 | 270 | 280 | 50 | 60 | 70 | 170 | 180 | 190 | 290 | 300 | 310 |
| 6 | 30 | 40 | 50 | 150 | 160 | 170 | 270 | 280 | 290 | 60 | 70 | 80 | 180 | 190 | 200 | 300 | 310 | 320 |
| 7 | 40 | 50 | 60 | 160 | 170 | 180 | 280 | 290 | 300 | 70 | 80 | 90 | 190 | 200 | 210 | 310 | 320 | 330 |
| 8 | 50 | 60 | 70 | 170 | 180 | 190 | 290 | 300 | 310 | 80 | 90 | 100 | 200 | 210 | 220 | 320 | 330 | 340 |
| 9 | 60 | 70 | 80 | 180 | 190 | 200 | 300 | 310 | 320 | 90 | 100 | 110 | 210 | 220 | 230 | 330 | 340 | 350 |
| 10 | 70 | 80 | 90 | 190 | 200 | 210 | 310 | 320 | 330 | 100 | 110 | 120 | 220 | 230 | 240 | 340 | 350 | 360 |

**Figure 4 – Example of the process of reshaping the dataset.**

### 1.2.2 Linear regression

Linear regression is the simplest regression algorithm, in which we assign numerical coefficients to each of the input features and the output variable is predicted as a linear combination of these input features and their respective coefficients, or weights. Training for this method consists of minimizing the mean square error between the observed and predicted data for the outputs (PANDEY et al. 2020). Linear regression, since it is simple to implement and provides understandable relationships between the analyzed variables, is generally applied in a stage of data exploration, and in this work, it was not different from that. As we have several input features, our problem becomes a multidimensional linear regression, but the same principles are still valid.

To use this regression, we also observed some conditions, as described by Nielsen (2020), assuming that the time series have an approximately linear response (observing short time windows) and that the input variables do not remain constant and do not present perfect correlations with each other.

Other variants of the linear regression include regularization terms, which act as penalties for the regression coefficients. That is a method to prevent overfitting, reducing the importance of some of the input variables that could capture noise or outliers and hinder the generalization capabilities of the regressor.

Ridge regression adds a term in the equation of loss function (that will be minimized in the training phase), as we see below:

$$f_{ridge} = \sum_{i=1}^{n}[y_i - (w \cdot x_i + b)]^2 + \alpha \cdot \sum_{j=1}^{p} w_j^2. \qquad (4)$$

This term is weighted by a factor $\alpha$, that can be tuned to control the penalties of the regression coefficients. Using $\alpha=0$ the equation is reduced to the simple linear regression, while using large values can exaggeratedly shrink the coefficients (KALITA, 2024).

Lasso regression considers a slight modification of the previous equation, using absolute values instead of squared values in the penalty component:

$$f_{lasso} = \sum_{i=1}^{n}[y_i - (w \cdot x_i + b)]^2 + \alpha \cdot \sum_{j=1}^{p} |w_j|. \qquad (5)$$

This can make some of the coefficients equal to zero (that doesn't happen in ridge regression), what can increase generalization capabilities and provide better interpretability of the relations between input and output variables (KALITA, 2024).

In this work, we implement and evaluate these three variants of linear regression.

### 1.2.3 Neural networks

An Artificial Neural Network (ANN), in the original paradigm inspired by biological neural networks, consists of a set of processing units (also called neurons or nodes) designed to provide an output value within a given range based on the weighted sum of inputs and subsequent application of an activation function. With varied possibilities of connections and arrangements, these networks are able to "learn" from examples and can be used in a wide variety of problems, including classification, approximation of functions, interpolations, etc. (BRAGA; CARVALHO; LUDERMIR, 2000). The most widespread model of Neural Network is the one known as MLP (acronym for Multilayer Perceptron), illustrated in Figure 5. This model is based on a direct signal propagation, with neurons arranged in sequential layers and the outputs of all neurons in each layer connected as inputs to the next layer. The knowledge of the network is stored in the weights associated with each of these connections (vectors $W_i$, $W_h$ and $W_o$ in the illustrated example), and learning is usually performed in iterative processes of adjusting these weights based on reference input-output relationships known *a priori*.

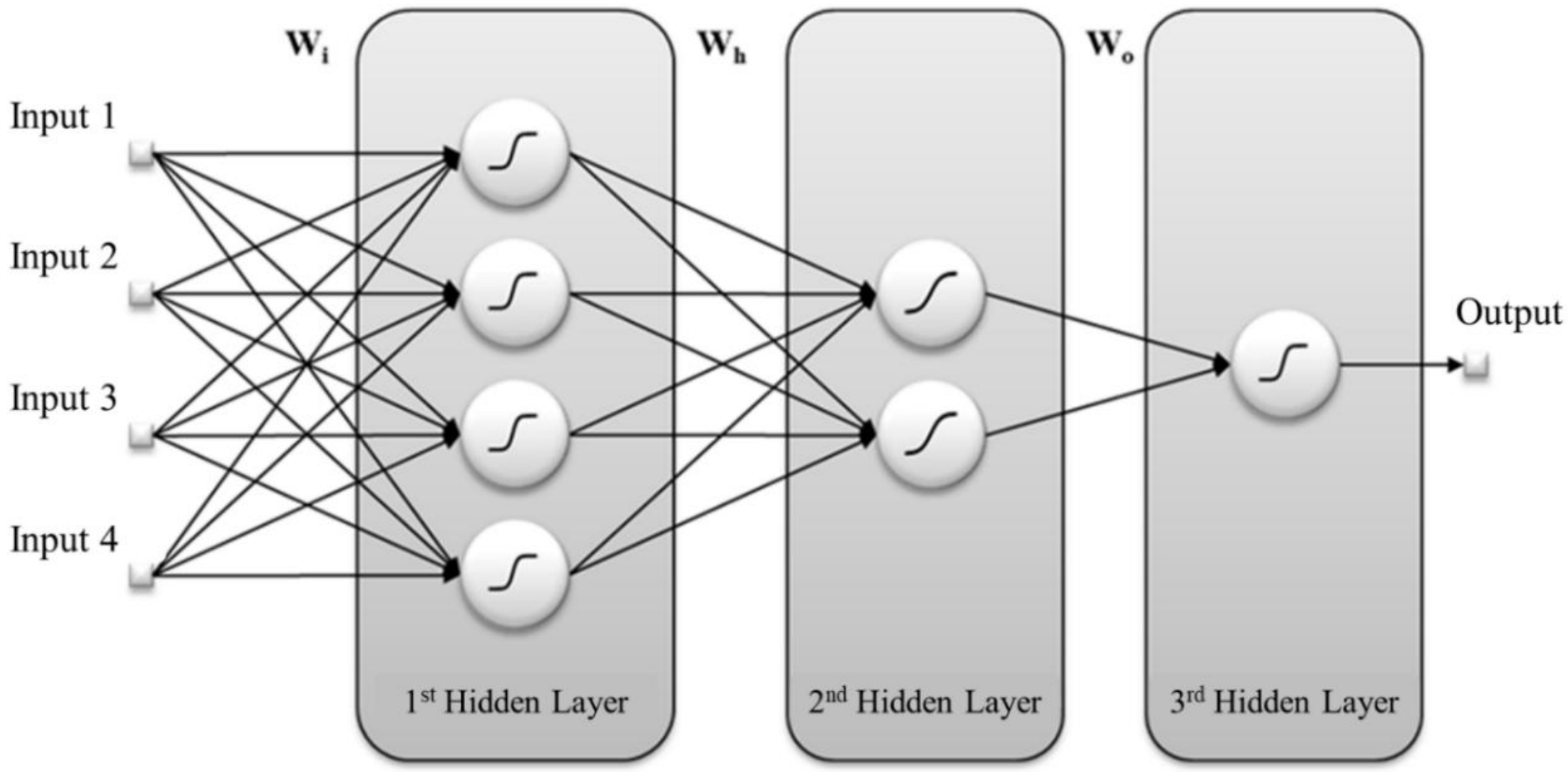

**Figure 5 – Example of the architecture of a Multilayer Perceptron Network.**

Each of the neurons of an ANN has an activation function, responsible for calculating its output value as a function of the weighted sum of its inputs. Some common activations are based on sigmoid functions ("s"-shaped), due to their balance between linear and non-linear behavior, and for being differentiable at all points, which is important for the training stage (BRAGA *et al.*, 2000). Other activations include step and linear functions, the latter sometimes used as a piecewise function, *e.g.* the Rectified Linear Unit (ReLU).

Regarding the training of an ANN, the most popular algorithm is the backpropagation. In this method, the weights of the connections between the nodes are initialized with random values. Then, sets of input values from the training dataset are inserted in the network. For each of these sets, the network outputs with the current weights are computed, in the forward phase of the training. The obtained output is then compared to the correct one calculating the error between both. This error is reversely propagated through the network (backward phase, hence the name of the algorithm). The product of the error of each output by a constant "learning rate" is subtracted from the connection weights of the respective node in the last layer. The error of each node of the previous layers is calculated using the errors of the nodes of the next layer connected to it, weighted by the weights of the connections. The procedure is repeated with new pairs of input / output vectors being presented to the network until a stop criterion is reached: the mean quadratic error becomes less than a pre-established limit; a maximum number of iterations is performed, or the error is stagnant between iterations. Other algorithms can be found in the literature (HAYKIN, 2008), sharing the idea that the adjustment of the weights makes the network capable of reproducing with the least possible divergence the input and output relations presented as a training set, with posterior validation and confirmation of its suitability for use with new data.

Recently, with the increase in the processing capabilities of CPUs and GPUs, it became easier to work with networks composed by multiple layers and a large number of neurons – the so-called Deep Learning. Those networks have a higher capability to generalize even in problems with non-linearities and high complexities (HAYKIN, 2008).

Another example of recent development is the family of the recurrent networks, as the Long Short-Term Memory (LSTM). Those networks operate with internal states in the neurons that can retain previous states, what make them more apt to deal with sequential data, as time series (ZHANG, 2021).

In this work, we use at first the simple MLP Neural Networks, and at the end we discuss how the latest advances in machine learning can bring improvements to our forecasts.

### 1.2.4 Evaluation metrics

We give special attention to suitable metrics for evaluating the performance of predictors/estimators while working with time series. In the literature we most often find measurements using the Mean Absolute Percentage Error (MAPE) and the Root Mean Squared Error (RMSE). These metrics have limitations such as the greater penalty for models that tend to diverge towards higher values (as the error can be unlimited in this direction but is limited when it is towards lower values). Therefore, we also include in the evaluations the Symmetric Mean Absolute Percentage Error (SMAPE), which will correct the problem described, despite showing a small tendency in the opposite direction. (LEWINSON, 2020). Another common way to quantify the quality of the fit is using the coefficient of determination $R^2$, that is intended to describe the proportion of variation in the calculated variable that is explained by our regression model (KALITA et al., 2024).

These metrics are given by the following equations:

$$MAPE = \frac{1}{n}\sum_{j=1}^{n}\left|\frac{y_j - \hat{y}_j}{y_j}\right| \quad (6)$$

$$RMSE = \sqrt{\frac{1}{n}\sum_{j=1}^{n}\left(y_j - \hat{y}_j\right)^2} \quad (7)$$

$$SMAPE = \frac{1}{n}\sum_{j=1}^{n}\frac{|y_j - \hat{y}_j|}{\frac{(y_j + \hat{y}_j)}{2}} \quad (8)$$

$$R^2 = 1 - \frac{\sum_{j=1}^{n}\left(y_j - \hat{y}_j\right)^2}{\sum_{j=1}^{n}\left(y_j - \bar{y}_j\right)^2} \quad (9)$$

where $y_j$ are the predicted values and $\hat{y}_j$ the actual values of the dependent variable (REZENDE, 2018).

## 1.3 Case Studies

Next, we present two case studies where we evaluate the proposed methodology: the first based on a simulated dataset and the second based on a real reservoir located in the Brazilian pre-salt.

### 1.3.1 Simulated dataset

Our first case study is based on synthetic production curves generated from the UNISIM III compositional simulation model (CORREIA et al., 2020). It is a model of a karst carbonate reservoir, created by researchers from University of Campinas (UNICAMP) with public data from the Brazilian pre-salt reservoirs complemented with synthetic data and made available as a public domain for use as a benchmark for academic purposes.

The version used, illustrated in Figure 6, represents an initial stage of development of the field, with wells concentrated in the top of the structure and distributed as follows:

- 06 oil producing wells - P11, P12, P13, P14, P15 and P16 (well P21, in an isolated area, will not be opened in the simulation).

- 07 injector wells, 03 initially configured for water injection (I12, I14 and I15) and the other 04 initially injecting gas (I11, I13, I16 and I17).

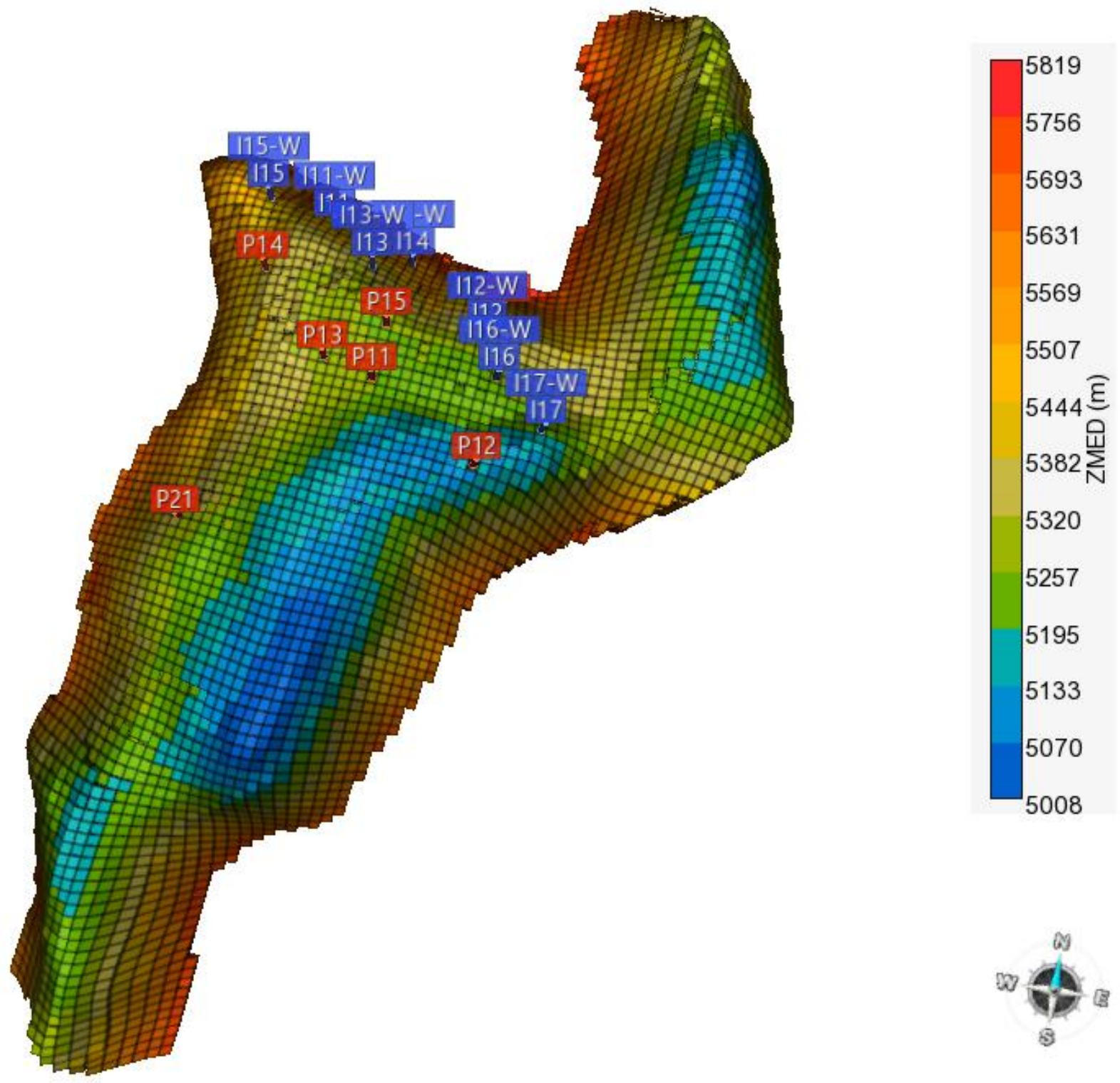

**Figure 6 – Representation of the reservoir simulation model UNISIM III (adapted from CORREIA et al., 2020).**

To generate a database for training and evaluation of the methods, we run a simulation of this model with extrapolation for a period of 20 years. An injection rate schedule was proposed with several changes in strategy over the simulation period, in an attempt to create richer data (greater variations over time) and replicate a production history more relatable to a real-life scenario, that include various sources of inefficiencies and unavailabilities of wells and equipments. The simulated curves are illustrated in Figure 7. It is relevant to note that in this case we can obtain directly from the simulation the potential production, without operational factors or any kind of noise in data, thus, not requiring an additional step in data conditioning, as we will see later in the case study of a real oilfield.

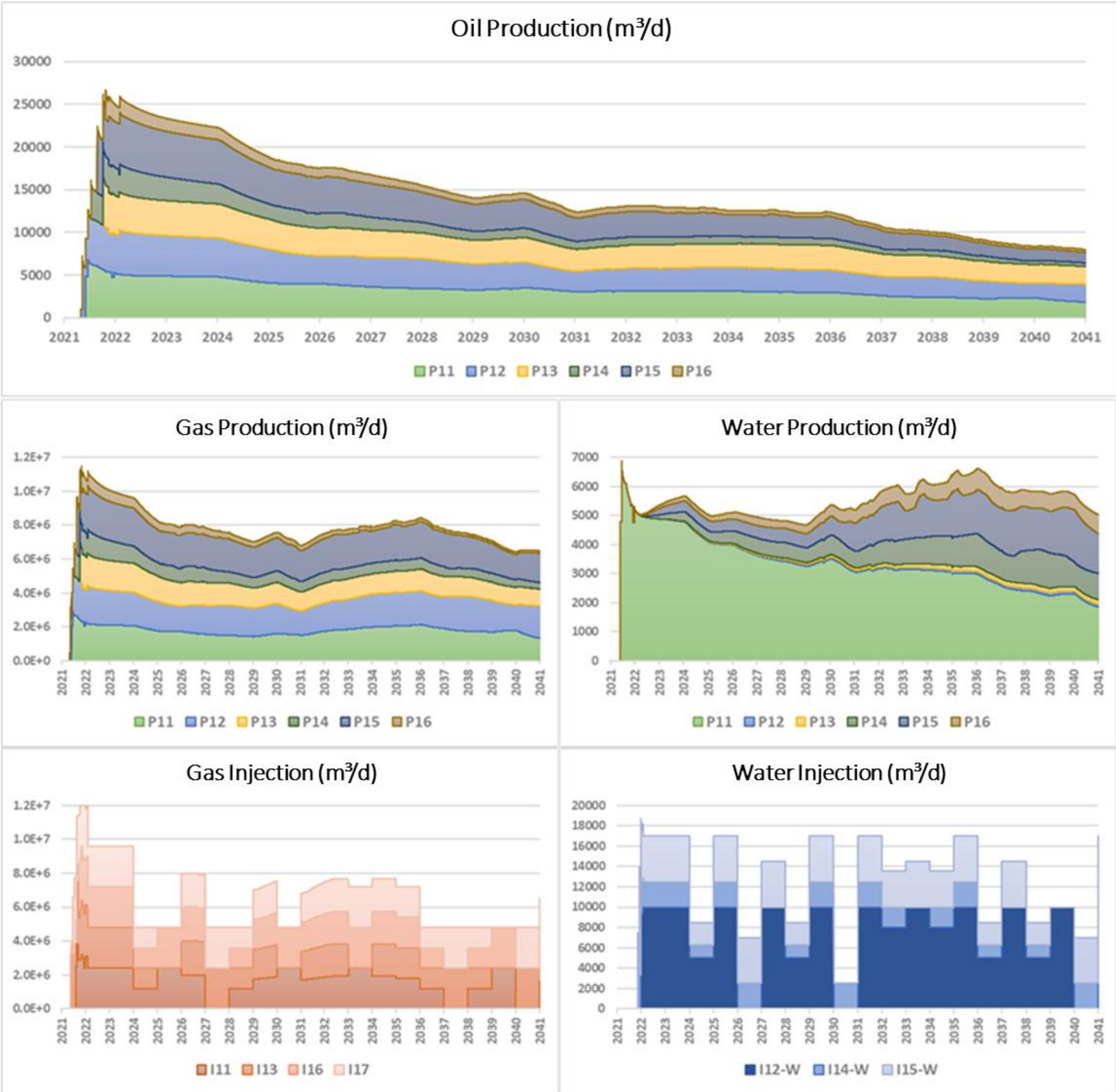

**Figure 7 – Simulation results for the UNISIM-III model as used in our case study.**

With this dataset, we explore and evaluate several configurations of estimators, and different approaches to the forecasting problem, including:

- Forecast for the entire field or segmented by well: Both approaches are important in practical cases, so we will evaluate the performance of the methods in these two variants.

- Data sampling (from originally daily rates): In this aspect we need a compromise solution, as sampling at short intervals allows greater granularity and more samples for the training set, but at the same time it implies in more forecast steps so that we have the desired horizon for practical management purposes, resulting in higher cumulative error.

- Single-step or step-by-step (incremental) forecast: In the first case, the predictor has as outputs from a single prediction step all the values for the defined look-forward window (according to Equation 3). In the second, the forecast outputs are given for a single future time step, but we use an iterative process, in which we calculate the next time step taking as inputs the $i$-1 past months of production/injection history (see Equation 2), the injection planned for the first future step (from a previously given injection schedule) and complementing it with the results of the production forecast from the previous step. This process is repeated until the desired production period is completed (at least 06 months, as we defined). It is important to note that the further we go into the future while forecasting, the more inputs will have true historical data replaced by previously forecasted data, which tends to amplify the cumulative error. However, in initial tests we identified a vastly superior performance of the step-by-step approach, especially due to the possibility of incorporating along-the-forecast injection schedules, and then we moved forward with this single option.

- Look-back and look-forward windows: Number of simultaneous steps to compose the inputs and outputs of the predictors. The look-back window will define the number of inputs for the estimator and, together with sampling, the time window captured, which is important because of the dynamics of the injections. As we chose the step-by-step approach, the look-forward in all subsequent evaluations will be $k$=1.

- Time intervals for training and retraining: Here we evaluate what would be the minimum history time to be used in training so that we have satisfactory performance and what is the ideal periodicity to retrain the estimator – incorporating new history data – and whether new training should be performed using incremental history or fixed windows, in this last one “forgetting” older history. This is important especially because we are dealing with non-stationary time series (since production rates generally tend to decrease over time) that are susceptible to concept drift (such as new water and/or gas breakthroughs).

- Machine learning methods, configurations, and hyperparameters: We evaluate variations of linear regressions and neural networks. For the regressions (Lasso and Ridge), we only have as a parameter the weight of the regularization term, α. For Neural Networks, however, the choice of architecture is fundamental for performance, being based on the complexity of the problem and the availability of data. Although it is difficult to define at a first glance the ideal network configuration for a given dataset, there are some references to the starting point in the evaluation, for example starting with the simplest possible network and gradually increasing its complexity (RUBO et al., 2019) or using empirical equations based on the numbers of inputs, outputs and samples, considering the undesired possibility of overfitting (AL-BULUSHI et al., 2012). In this work we try to reconcile these recommendations while choosing the number of elements in a single hidden layer. We tried different activation functions, and the training algorithm was the *Adam* (SCIKIT LEARN, 2020).

Initially, we evaluated the production forecasts for the entire field, without individualization per well. Thus, the variables are the total production rates $Q_o$, $Q_g$ and $Q_w$, and the injection rates $Q_{wi}$ and $Q_{gi}$, forming databases as described in Section 1.2.1

The evaluation we conducted considers a grid search for the previously described parameters, not only observing the optimal combination, but also their individual behaviors and cross-relationships.

Evaluating for look-back and sampling windows, we calculated metrics from a variety of predictors, using incremental training set and calculated the average values; we preferably consider the SMAPE and MAPE metrics (shown in the plots), as they are more suited to time series, but we also observed the behavior of R2, MAE, and MSE. We established some limits for the observation windows in the analysis, considering the time necessary to capture the reservoir dynamics. This time is obtained by multiplying the number of look-back samples and the sampling period. From the plot in Figure 8, we notice that the combination of 5-day sampling and 15-samples look-back results in the best values, on average. However, we selected the combination of 10 days and 15 samples for the sequence of the analysis, so we have a longer time frame to capture events and give more stability in step-by-step forecasts.

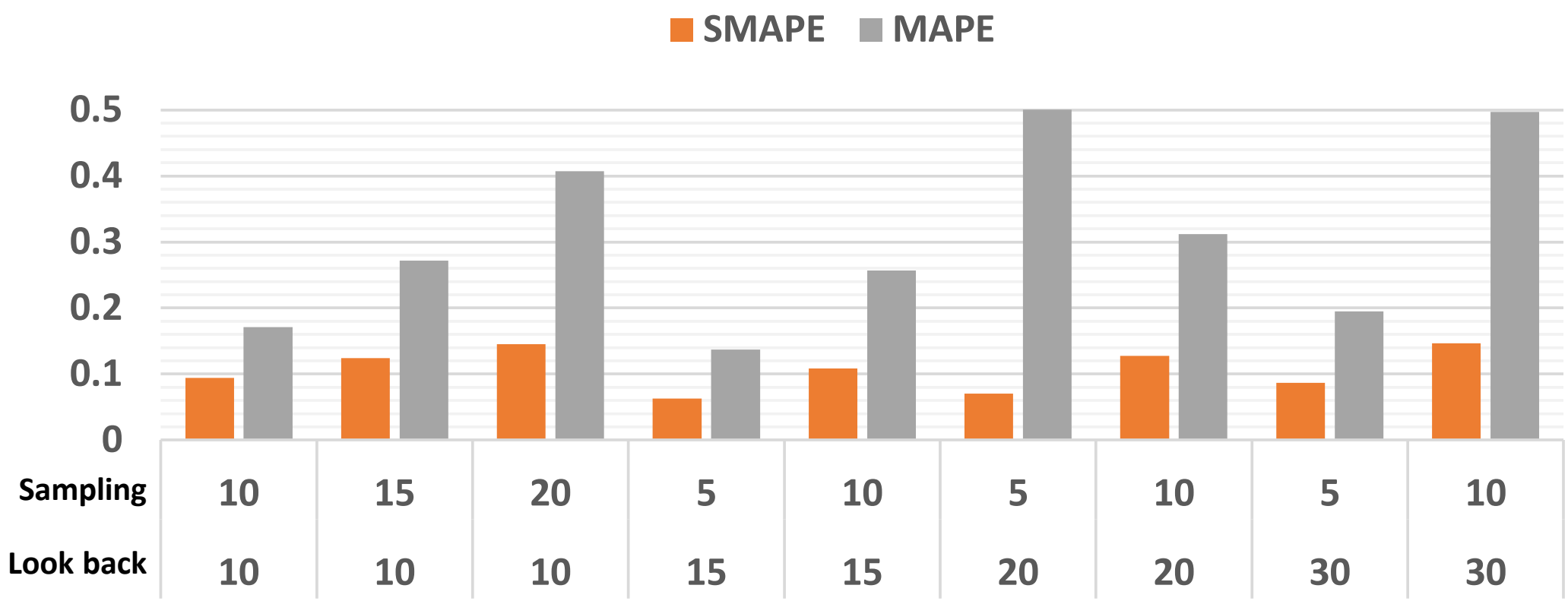

**Figure 8 – Evaluation metrics for the parameters sampling and look-back, for the case study of a simulated full field.**

In the next analysis, the plot in Figure 9 shows the error metrics as a function of the type of estimator and the minimum training window. Except for some noise, the general trend is that the longer the window, the better the accuracy, so we recommend working with incremental windows, that is, training with all the history available at the starting moment of the forecasts, not discarding older data after "aging".

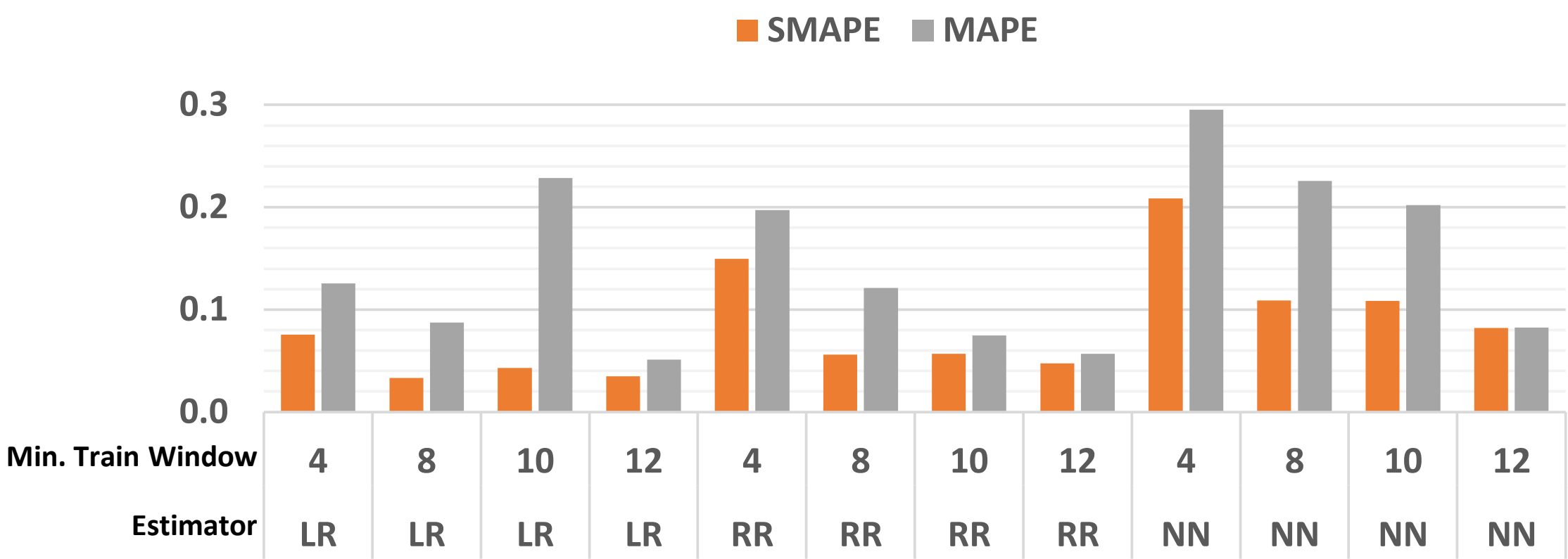

**Figure 9 – Evaluation metrics for the parameters training window and type of estimator, for the case study of a simulated full field.**

For the retraining, we found that when made after shorter periods the results are better. The choice here is to retrain yearly.

When evaluating the estimators, we considered several possible configurations. Table 2 presents a compilation of the evaluation metrics for the most relevant choices: simple linear regression, ridge regression (that was systematically superior to Lasso regression, which is why we did not include the latter in the compilation of this case) with different alpha values and neural networks with different options for the number of elements in the hidden layer.

**Table 2 – Compilation of evaluation metrics for selected estimators. Case study of a simulated full field.**

| Metric | LR | RR | | | NN | | |
|---|---|---|---|---|---|---|---|
| | - | α=0.05 | α=0.20 | α=0.40 | 10 el. | 20 el. | 70 el. |
| SMAPE | 0.105 | 0.116 | 0.125 | 0.128 | 0.149 | 0.146 | 0.148 |
| MAPE | 0.298 | 0.328 | 0.301 | 0.325 | 0.472 | 0.367 | 0.413 |
| MAE | 0.027 | 0.033 | 0.041 | 0.046 | 0.086 | 0.074 | 0.072 |
| MSE | 0.002 | 0.002 | 0.003 | 0.004 | 0.015 | 0.010 | 0.010 |

For a more intuitive visualization, but without quantitative rigor, we present a comparison of these metrics in a radar plot (Figure 10). In this plot, we use normalized values relative to the highest value of each of the metrics, that is, the outer vertices represent the worst performance and the inner represent the lowest error values. We can observe that the simplest method, linear regression, outperformed the other methods in this case. Although it can be considered as a surprise, this can be explained by the fact that the total field curves normally have smoother variations, and can be represented at a certain point by linear relationships, as pointed out by KUBOTA (2019).

From these results, we selected as models for the next evaluations the simple linear regression, the ridge regression with $\alpha$=0.2 and the neural networks with 20 elements in the hidden layer. The activation function used was *identity* (grid search showed that others, such as *ReLU* or *tanh* are less efficient in this problem).

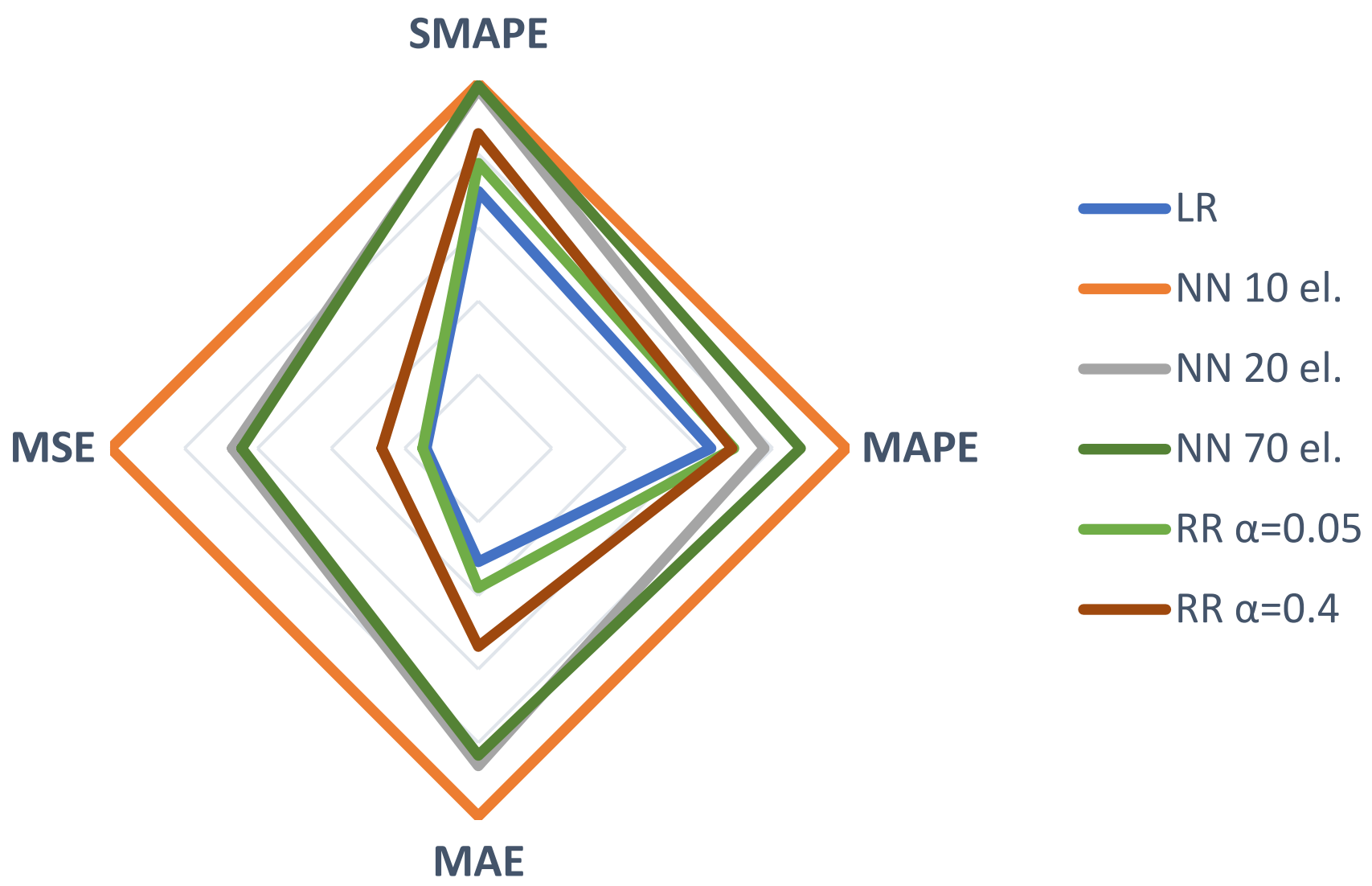

**Figure 10 – Radar plot comparing qualitatively the evaluation metrics for the selected estimators. Case study of a simulated full field.**

A more adequate way to observe the performance of the methods is to perform forecasts for an observation period and compare them with the reference values, in this case, those obtained from the simulation model. Such comparison is presented in the following examples.

In the first, we selected as $t$=0 a moment of constant gas injection and a couple days after a step up in the water injection rates, as shown in Figure 11. The forecast results for the next 06 months (18 steps, 10-day sampling) are presented in the sequence, for the Neural Network with 20 neurons (Figure 12), the Ridge regression with $\alpha$=0.2 (Figure 13), and the linear regression (Figure 14).

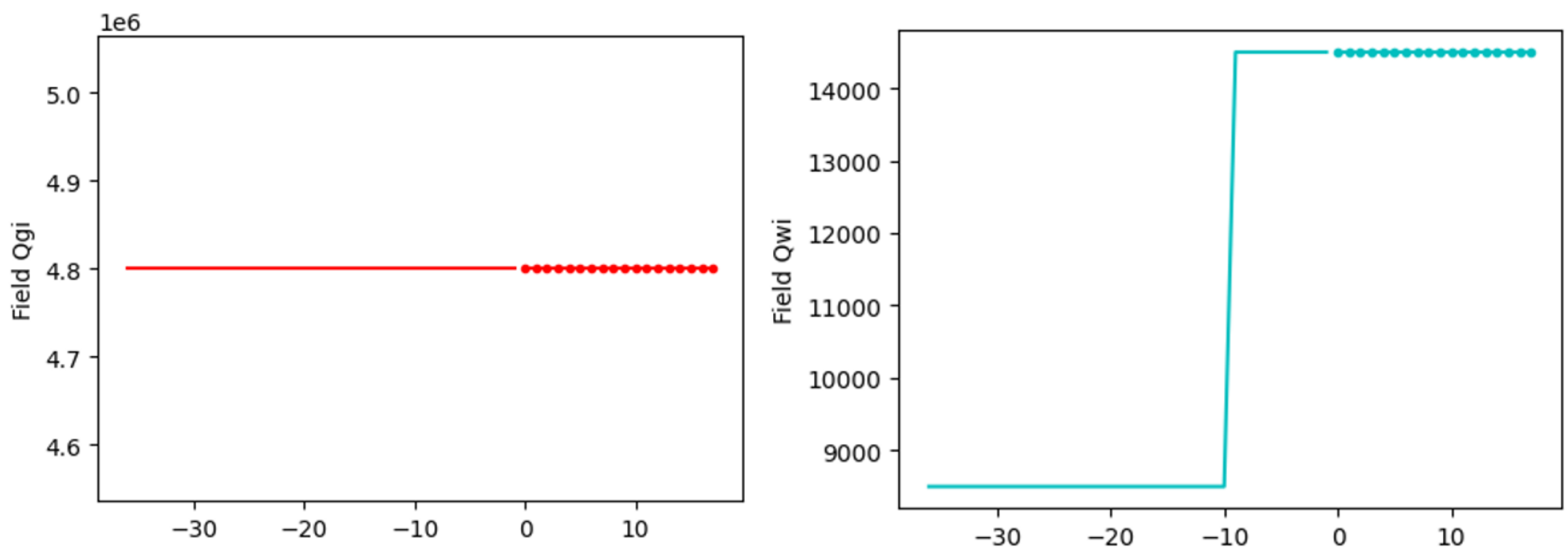

**Figure 11 – Injection schedule in the first example for the full field simulated case. Values in m³/d.**

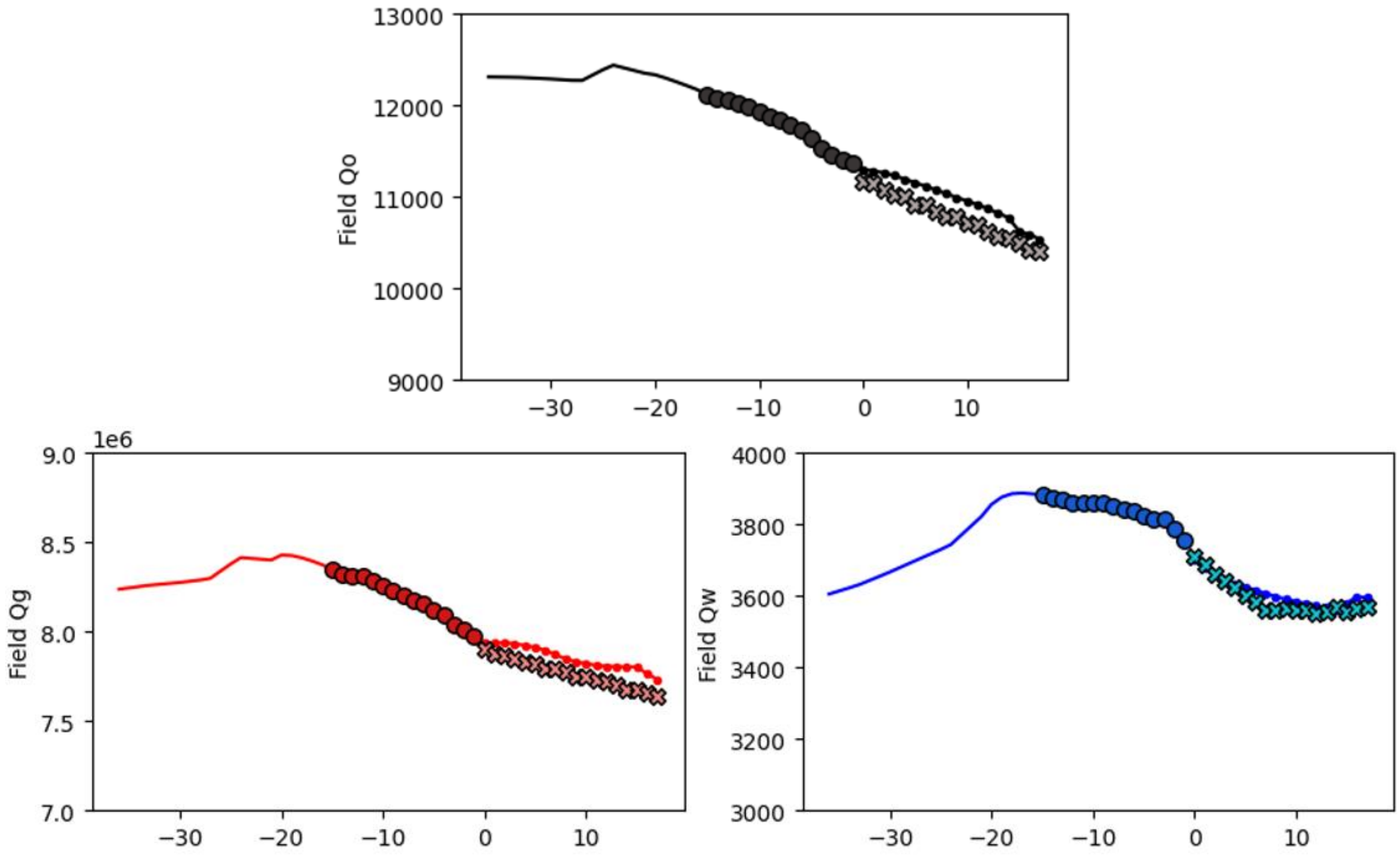

**Figure 12 – Oil, Gas and Water production for the first example for the full field simulated case. Forecast performed by Neural Network for the next 180 days (X on the curve), using as inputs the previous 150 days (circles on the curve). Values in m³/d.**

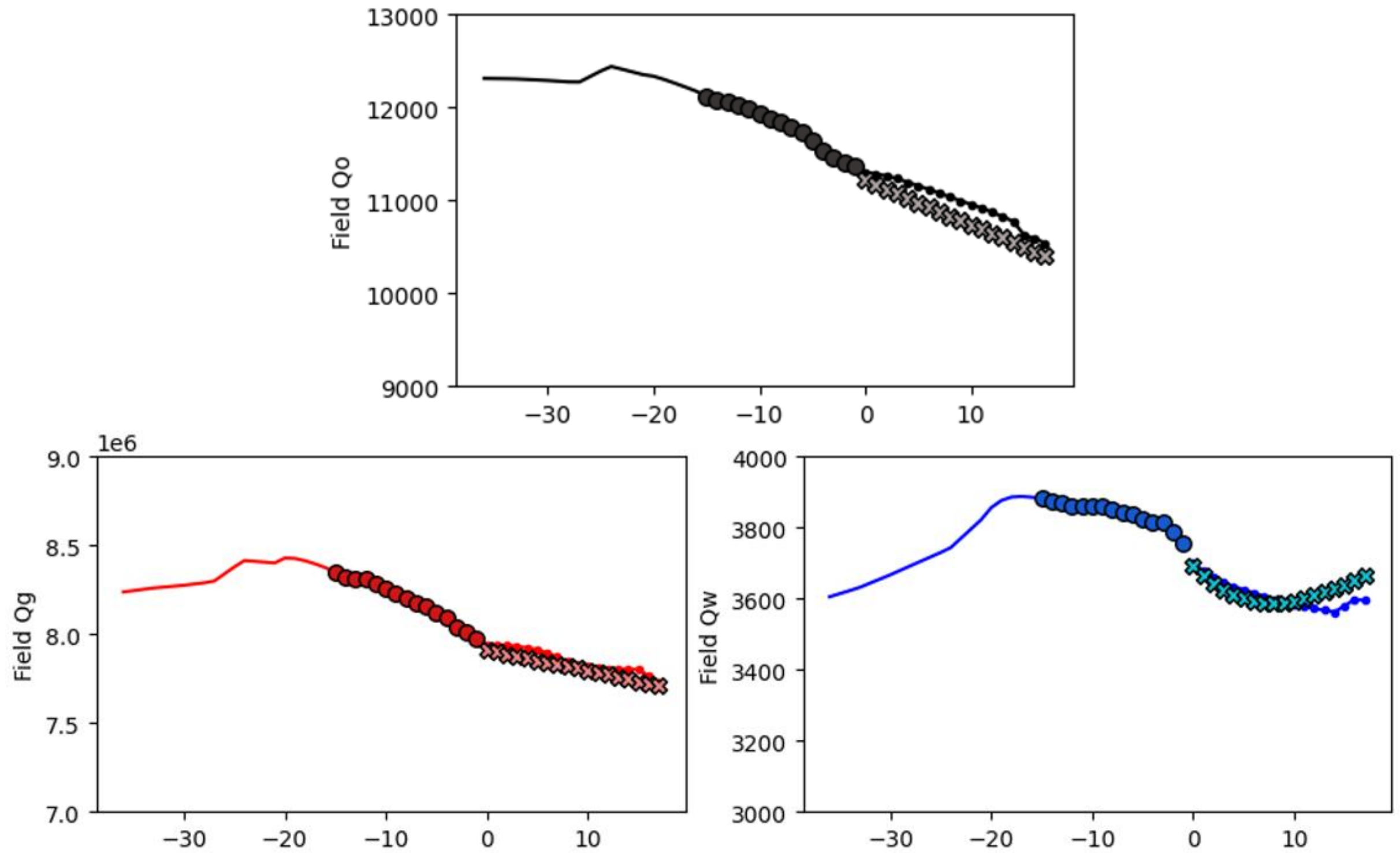

**Figure 13 – Oil, Gas and Water production for the first example for the full field simulated case. Forecast performed by Ridge Regression for the next 180 days (X on the curve), using as inputs the previous 150 days (circles on the curve). Values in m³/d.**

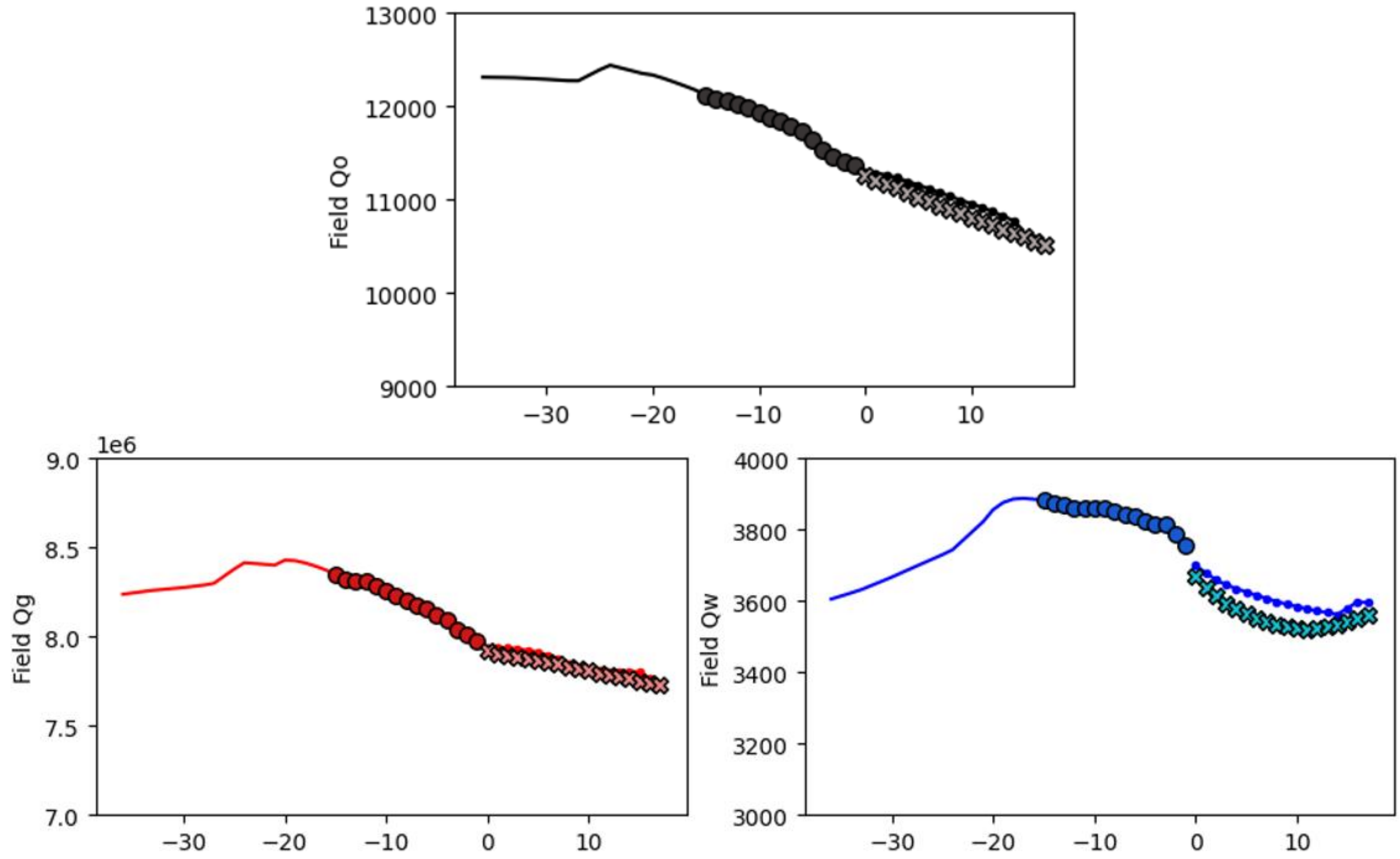

**Figure 14 – Oil, Gas and Water production for the first example for the full field simulated case. Forecast performed by Linear Regression for 180 days (X on the curve), using as inputs the previous 150 days (circles on the curve). Values in m³/d.**

For the second example, we selected a more complex moment regarding the injection schedules, as shown in Figure 15. The forecast results for the next 06 months (18 steps, 10-day sampling) are presented in the sequence, for the Neural Network with 20 neurons (Figure 16), the Ridge regression with $\alpha$=0.2 (Figure 17), and the linear regression (Figure 18)

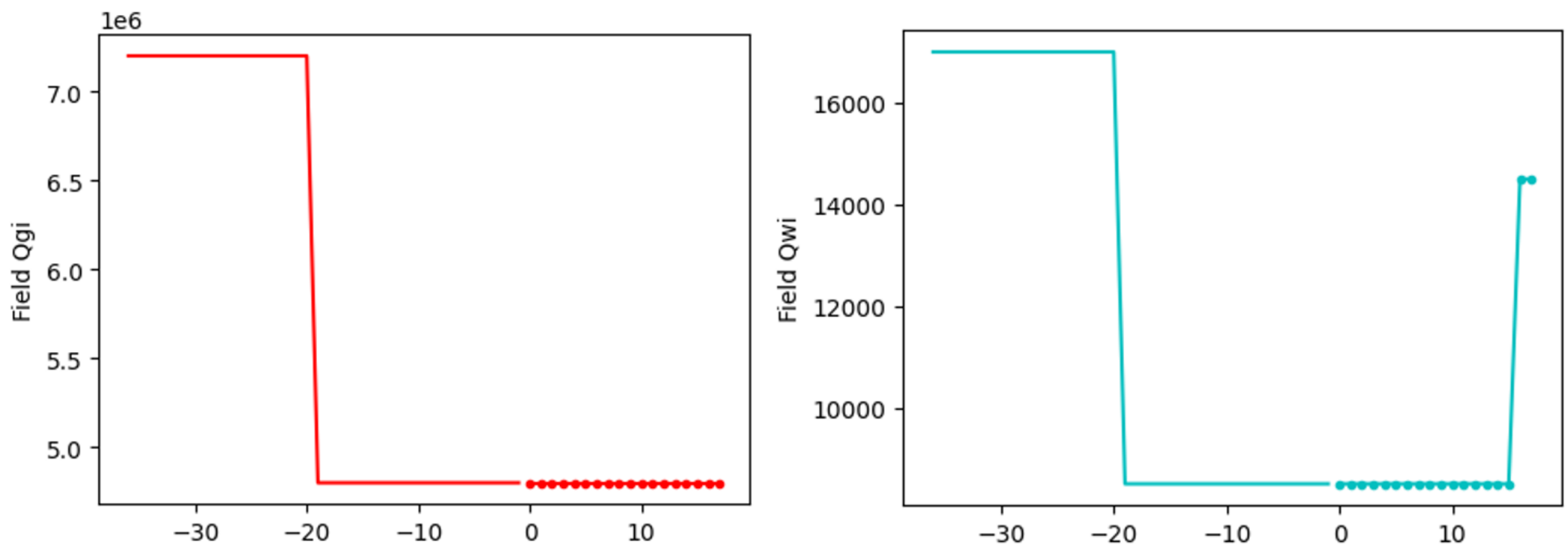

**Figure 15 – Injection schedule in the second example for the full field simulated case. Values in m³/d.**

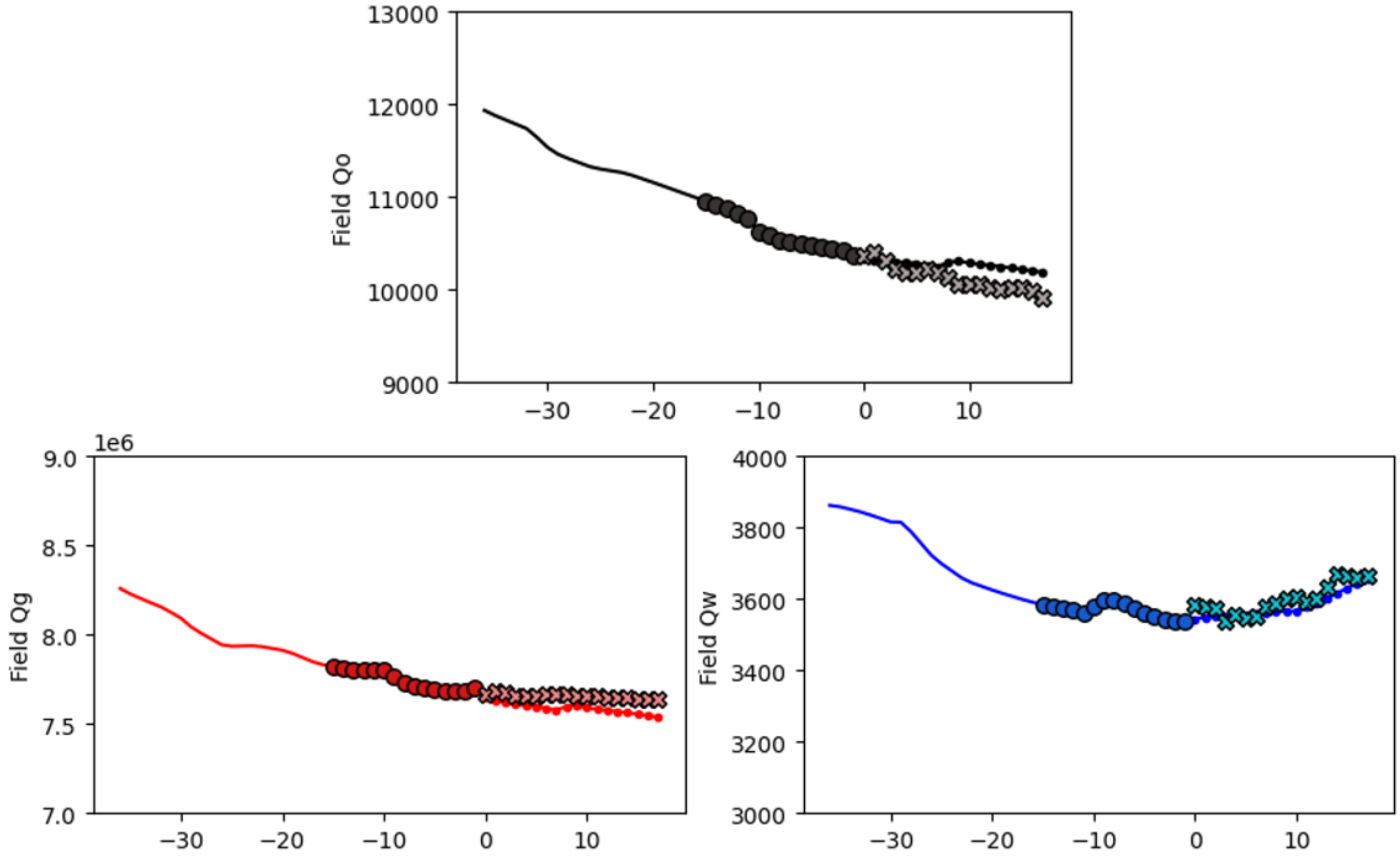

**Figure 16 – Oil, Gas and Water production for the second example for the full field simulated case. Forecast performed by Neural Network for 180 days (X on the curve), using as inputs the previous 150 days (circles on the curve). Values in m³/d.**

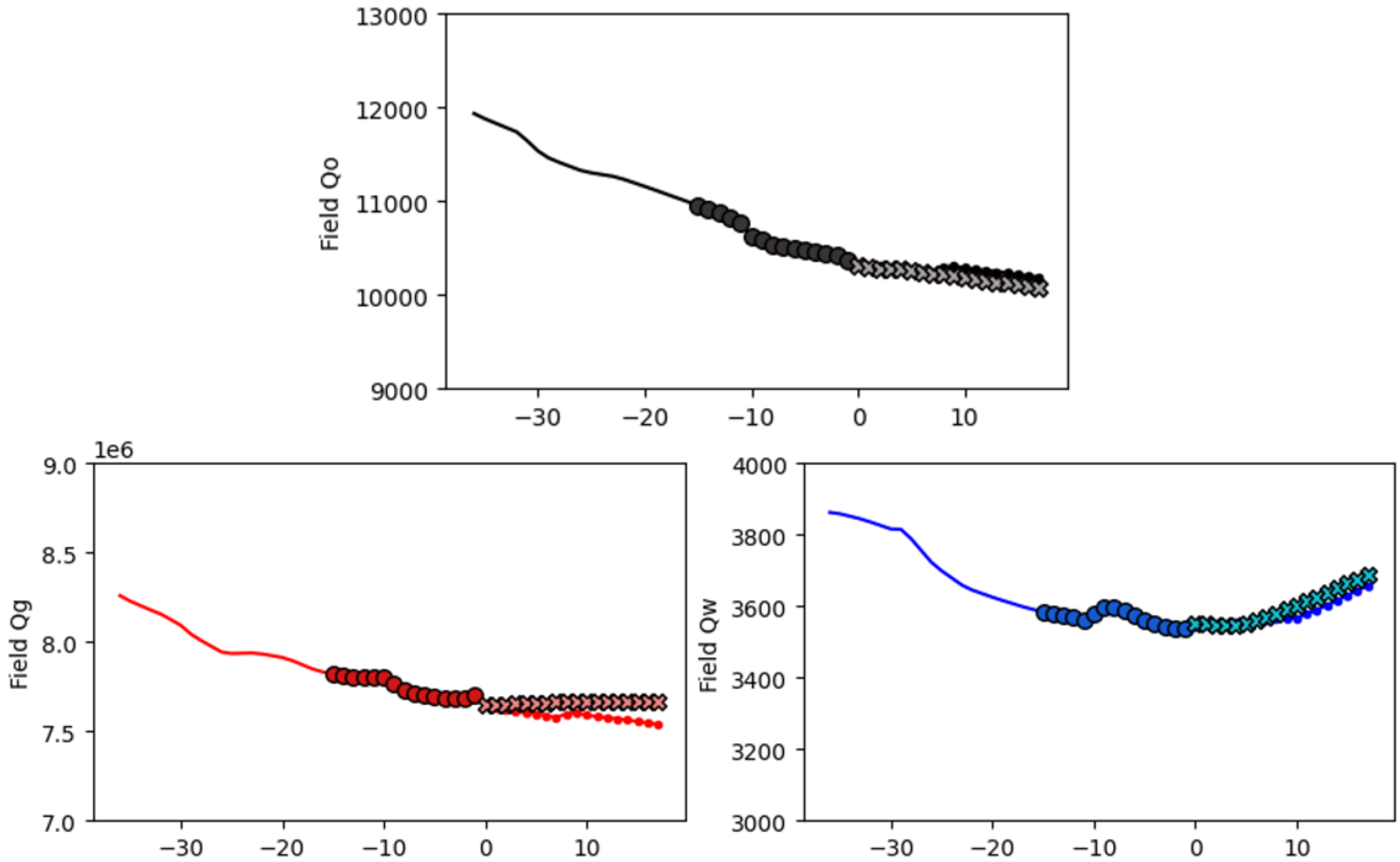

**Figure 17 – Oil, Gas and Water production for the second example for the full field simulated case. Forecast performed by Ridge Regression for 180 days (X on the curve), using as inputs the previous 150 days (circles on the curve). Values in m³/d.**

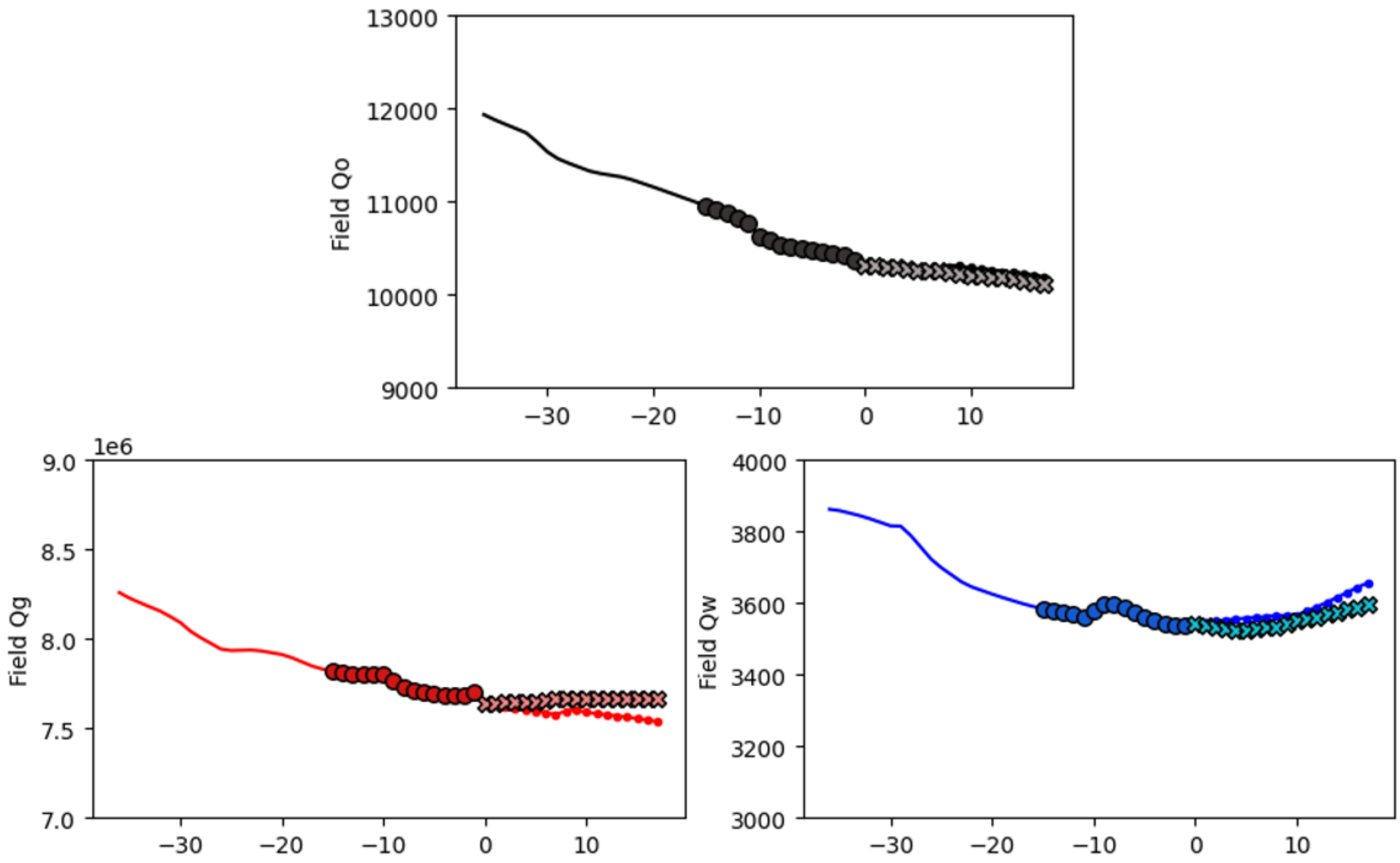

**Figure 18 – Oil, Gas and Water production for the second example for the full field simulated case. Forecast performed by Linear Regression for 180 days (X on the curve), using as inputs the previous 150 days (circles on the curve). Values in m³/d.**

In the next stage, we evaluate the methodology in a case study segmented per well. We start from estimator configured as the results of the study from the previous case.

Here we also observe, in Figure 19, that a longer training history tends to give better results (note that incremental training has a "burden" of starting with a shorter period, which is why it loses in metrics compared to the minimum training options of 10 or 12 years). Similarly to the previous case, here we also searched for the best choice for the sampling and look-back combination, obtaining this time the values of 10 days and 25 samples, respectively.

Evaluating the estimators, in this case, the linear regression did not present adequate results, perhaps due to the greater complexity in the relationships among wells, with more prominent non-linearities. Ridge regression was the best performing method across all metrics, as we can see in Table 3 and in the radar chart in Figure 20.

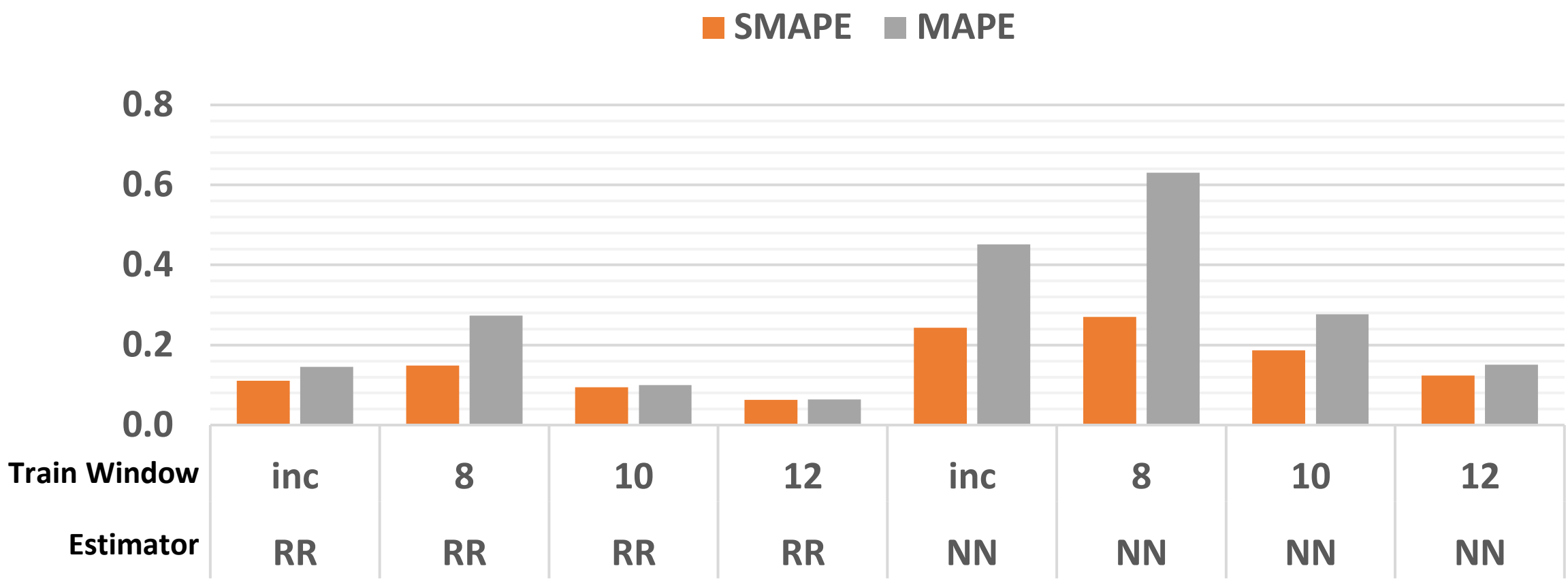

**Figure 19 – Evaluation metrics for the parameters training window and type of estimator, for the case study of an oilfield segmented per well.**

**Table 3 – Compilation of evaluation metrics for selected estimators. Case study of a simulated oilfield segmented per well.**

| Metric | LR | RR | | | NN | | |
|---|---|---|---|---|---|---|---|
| | α=0 | α=0.05 | α=0.20 | α=0.60 | 10 el. | 20 el. | 40 el. |
| SMAPE | 0.322 | 0.158 | 0.151 | 0.147 | 0.235 | 0.215 | 0.214 |
| MAPE | 1.446 | 0.186 | 0.272 | 0.182 | 0.558 | 0.444 | 0.364 |
| MAE | 0.806 | 0.196 | 0.179 | 0.176 | 0.286 | 0.265 | 0.264 |
| MSE | 6.900 | 0.108 | 0.081 | 0.072 | 0.163 | 0.144 | 0.157 |

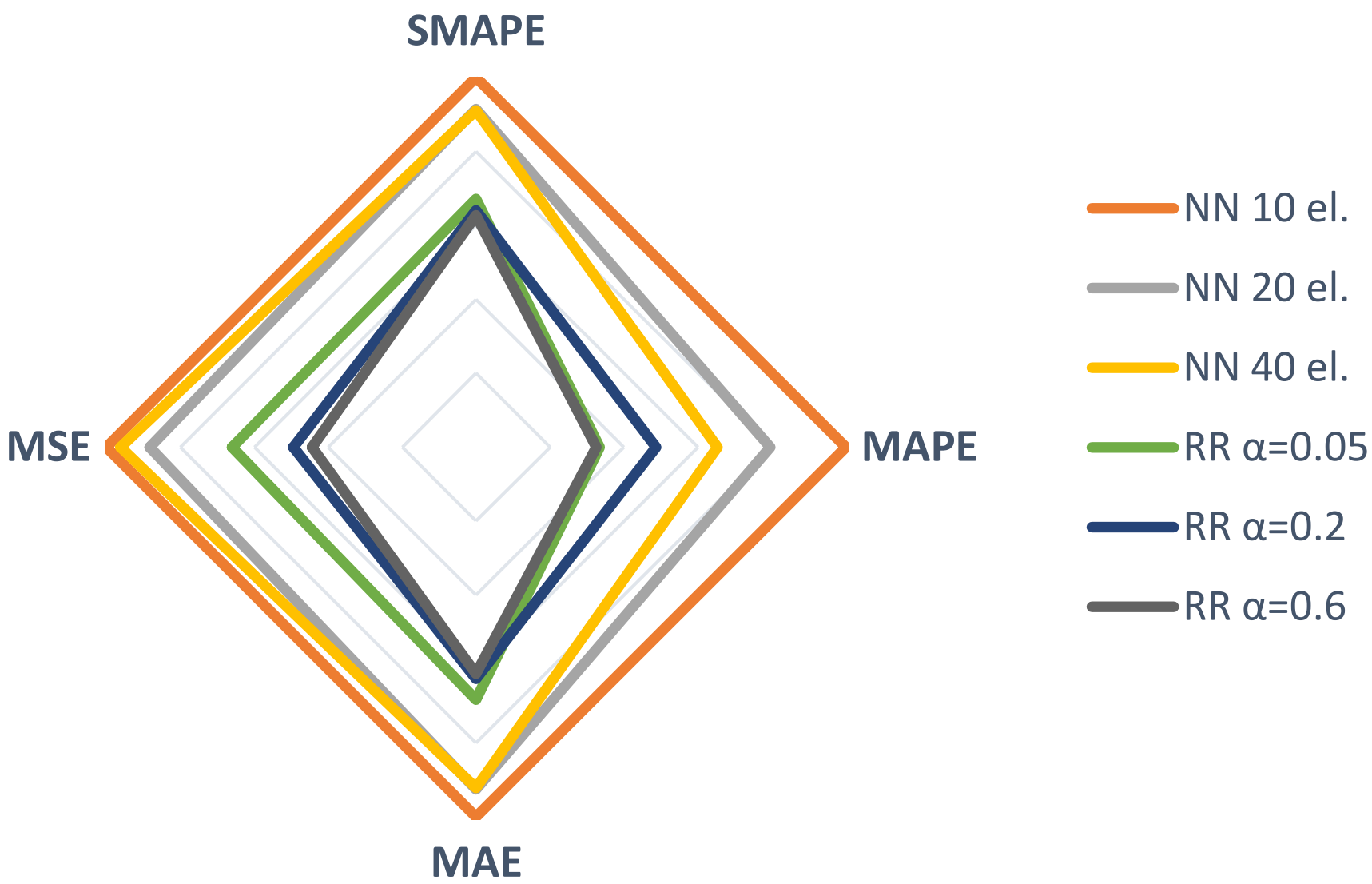

**Figure 20 – Radar plot comparing qualitatively the evaluation metrics for the selected estimators. Case study of a simulated oilfield segmented per well.**

Finally, we analyze the curves, this time forecasting a period of 12 months in the future. Injection schedule for this observation period is shown in Figure 21, and the results using our best estimator (Ridge regression with $\alpha$=0.6) are presented in Figure 22, for the 6 producing wells operating in the model. We can notice some divergences caused by the cumulative error due to the step-by-step forecast, but in general the estimator is capable of satisfactorily reproducing the trends.

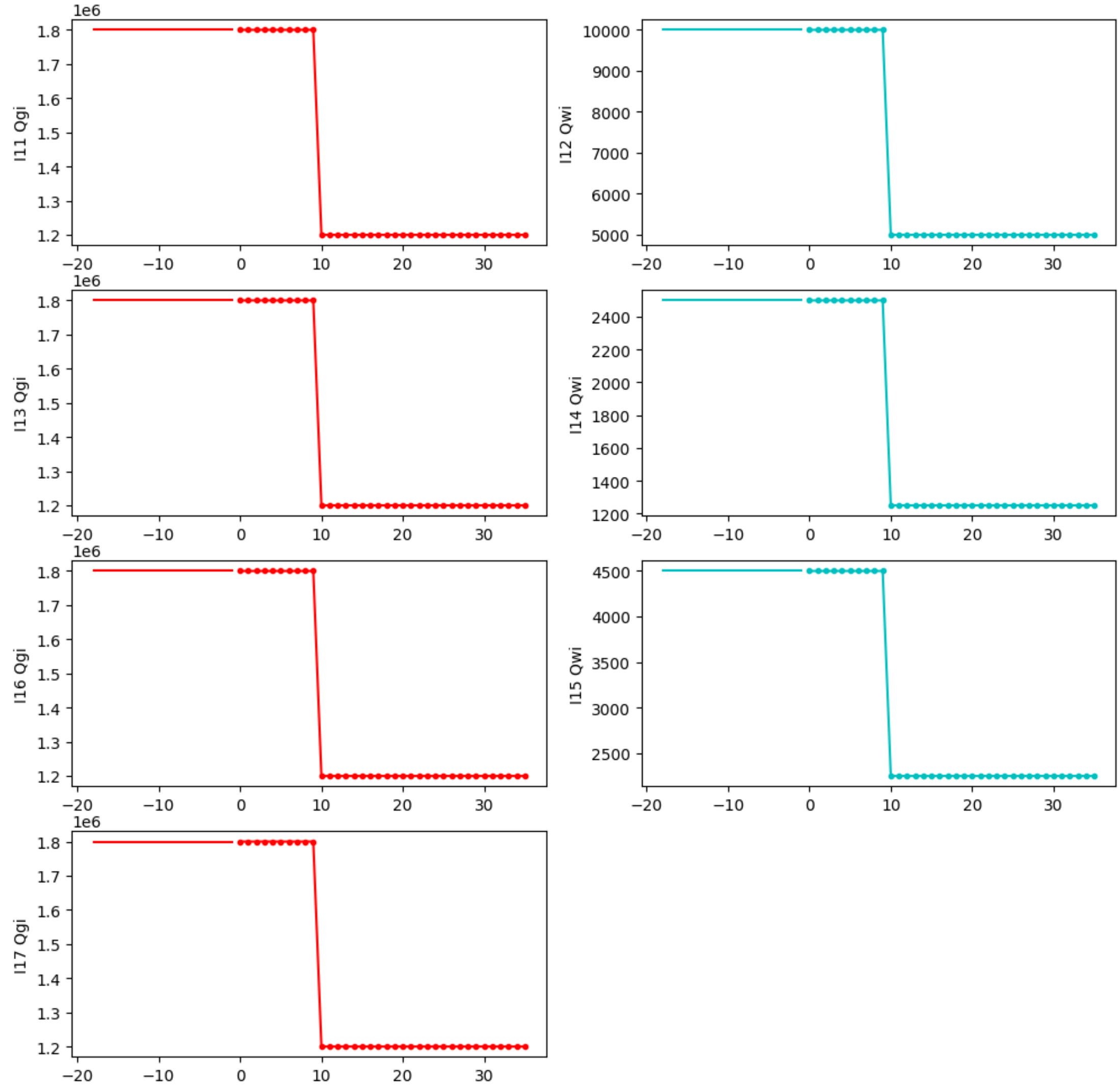

**Figure 21 – Injection schedules in the example of forecast segmented by well. Values in m³/d.**

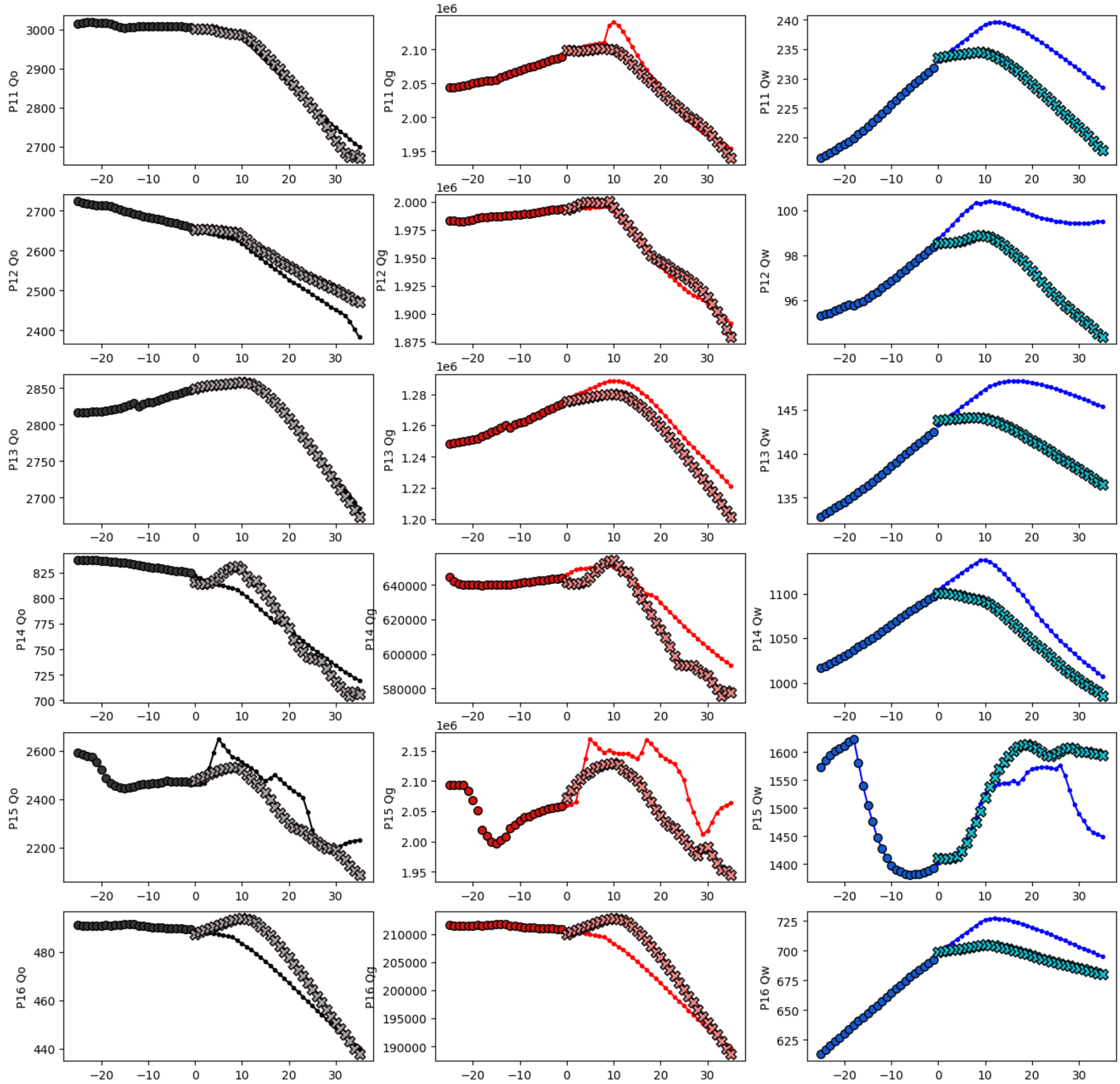

**Figure 22 – Oil, Gas and Water production for the example of forecast segmented by well. Forecast performed by Ridge Regression for 360 days (X on the curve), using as inputs the previous 250 days (circles on the curve). Values in m³/d.**

### 1.3.2 Pre-salt reservoir dataset

In this case study we applied the same framework to a dataset obtained from a giant offshore oilfield located in the Brazilian pre-salt. This is a carbonate reservoir, subjected to water alternating gas (WAG) for secondary recovery. Due to the high complexity of the reservoir and the relations among wells, considering the WAG cycles, miscible gas injection etc., we selected a relatively small part, considering that a more complete representation must require several complementary data-based models. In Figure 23 we show a geological map of the selected portion, focusing on the producer P4 and the influences of the surrounding wells, notably the WAG injectors I7 and I9. For confidentiality reasons we cannot disclose its data, so we adopted alternative names for these wells.

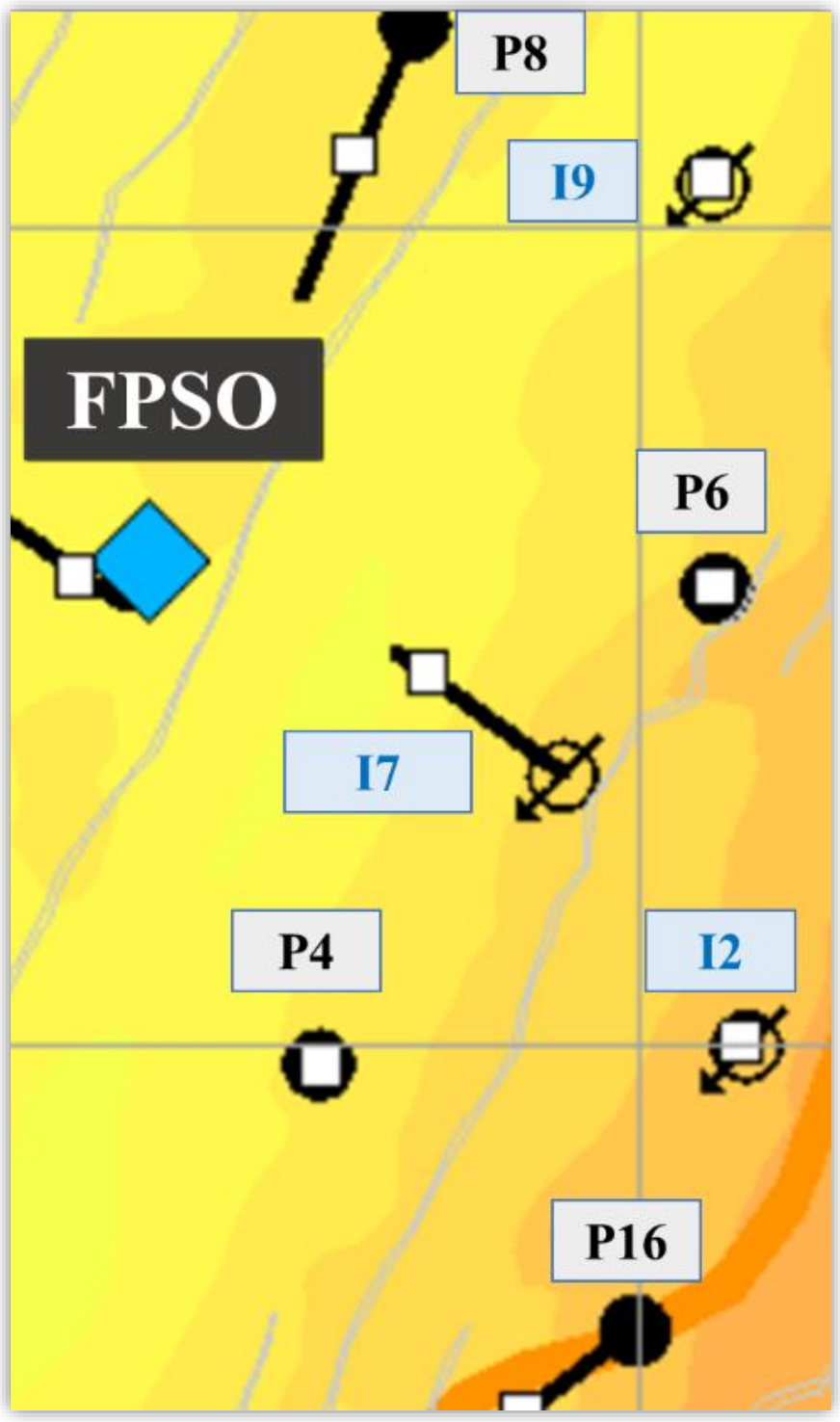

**Figure 23 – Map of the selected area from an oilfield in the Brazilian pre-salt.**

Similarly to the previous case studies, we start from obtaining our dataset. In this case, from the database of the operator company. In this case, however, we need to include an extra step in data conditioning, that is to remove "noise" from the time series. That comes from oscillating behavior, operational issues, temporary unavailability of wells or equipments, among other "real-world" influences.

Furthermore, we must deal with the fact that official production measurements on a platform are carried out as a whole, and not individually per well. The allocation of daily values segmented by well is based on a breakdown of total production, weighted according to the values from the most recent production test of each well in operation.

So, we estimate the production potential over time, using not only the daily rates, but also the results of production tests. We accomplished that by using a combination of linear interpolation and backward filling. The results for the oil, gas, and water rates from producer P4 are exhibited in Figure 24.

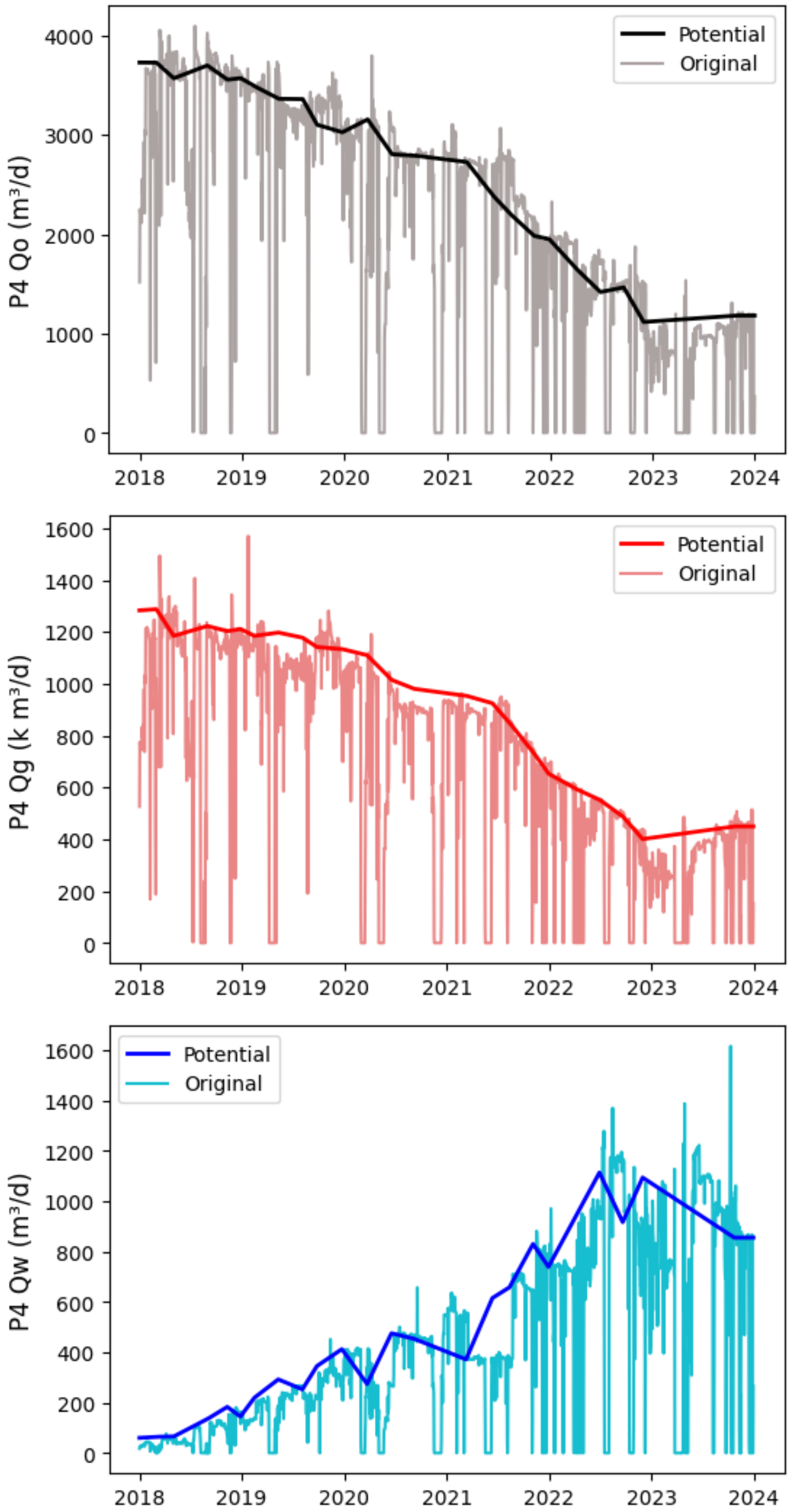

**Figure 24 – Oil, gas, and water rates from well P4, comparing the original curves and the potential obtained after data conditioning.**

For injection wells, the same noise generating situations apply. In this case, our smoothing method is calculating moving averages, considering that a delay will occur for the influence on the producer. The results for WAG injector I7 are shown in Figure 25.

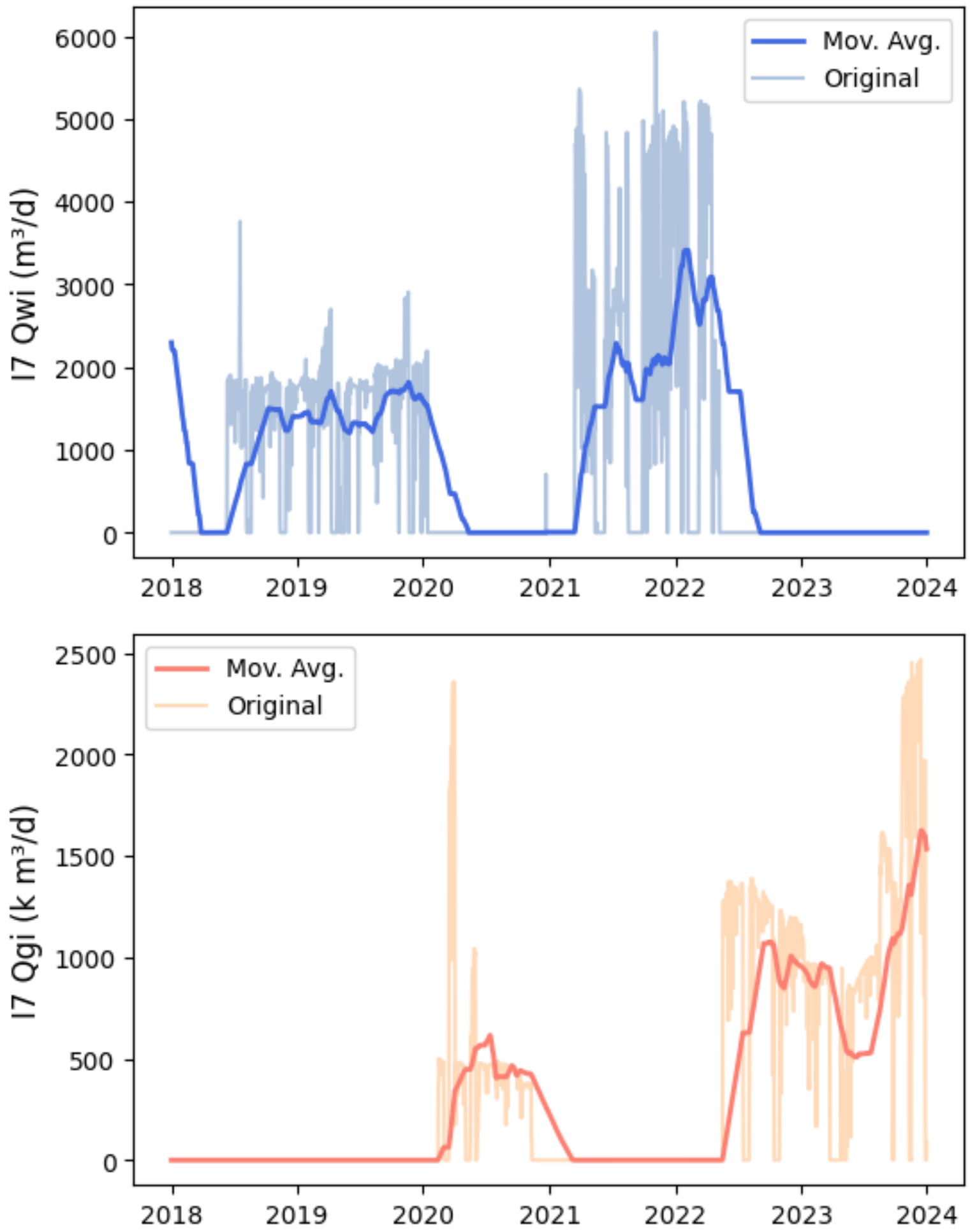

**Figure 25 – Gas and water injection rates for well I7, comparing the original curves and the results of smoothing using moving averages.**

The next steps in the analysis are the same as in the previous case studies, and again we analyzed by grid search the influences of several parameters in the performance. Preliminary results using validation dataset showed us that the influences from water injector I2 (partially isolated by a fault) and WAG injector P9 are not relevant, causing negative effect on the forecasts for P4. So, we used only information from WAG injector P7 alongside information from producer P4.

The sampling was defined as the 20-days average, and the look-back as 10 samples. The minimum training period is three years, retraining after each 06 months. The curves start in 2018, so the first validations start in 2021. At each training round we estimate production for the next 6 months, calculate the error metrics for this period and, subsequently, retrain incorporating the recent production/injection history.

The compilation of evaluation metrics for selected estimators is presented in Table 4 and in Figure 26. In this case, Lasso regression had a superior performance, but the best value for the regularization weight was not clear from these average metric values. That is because they were affected by some isolated cases of large errors, that were not excluded from the calculation. In a version of this framework to be made available to final users, a solution to mitigate these cases would be to allow manual fine-tuning of the forecasts, possibly excluding any visually noticeable outliers.

**Table 4 – Compilation of evaluation metrics for selected estimators. Case study of a single well in a real oilfield.**

| Metric | LR | | | RR | | | NN | | |
|---|---|---|---|---|---|---|---|---|---|
| | α=0.003 | α=0.05 | α=0.01 | α=0.1 | α=0.20 | α=0.40 | 20 el. | 40 el. | 70 el. |
| SMAPE | 0.346 | 0.355 | 0.314 | 0.493 | 0.490 | 0.483 | 0.576 | 0.607 | 0.558 |
| MAPE | 1.313 | 0.573 | 0.701 | 1.618 | 3.418 | 2.264 | 1.490 | 3.197 | 1.439 |
| MAE | 0.741 | 0.780 | 0.989 | 1.372 | 1.321 | 1.284 | 2.076 | 1.989 | 1.968 |
| MSE | 1.835 | 1.226 | 7.313 | 7.875 | 7.179 | 6.946 | 29.795 | 28.149 | 23.220 |

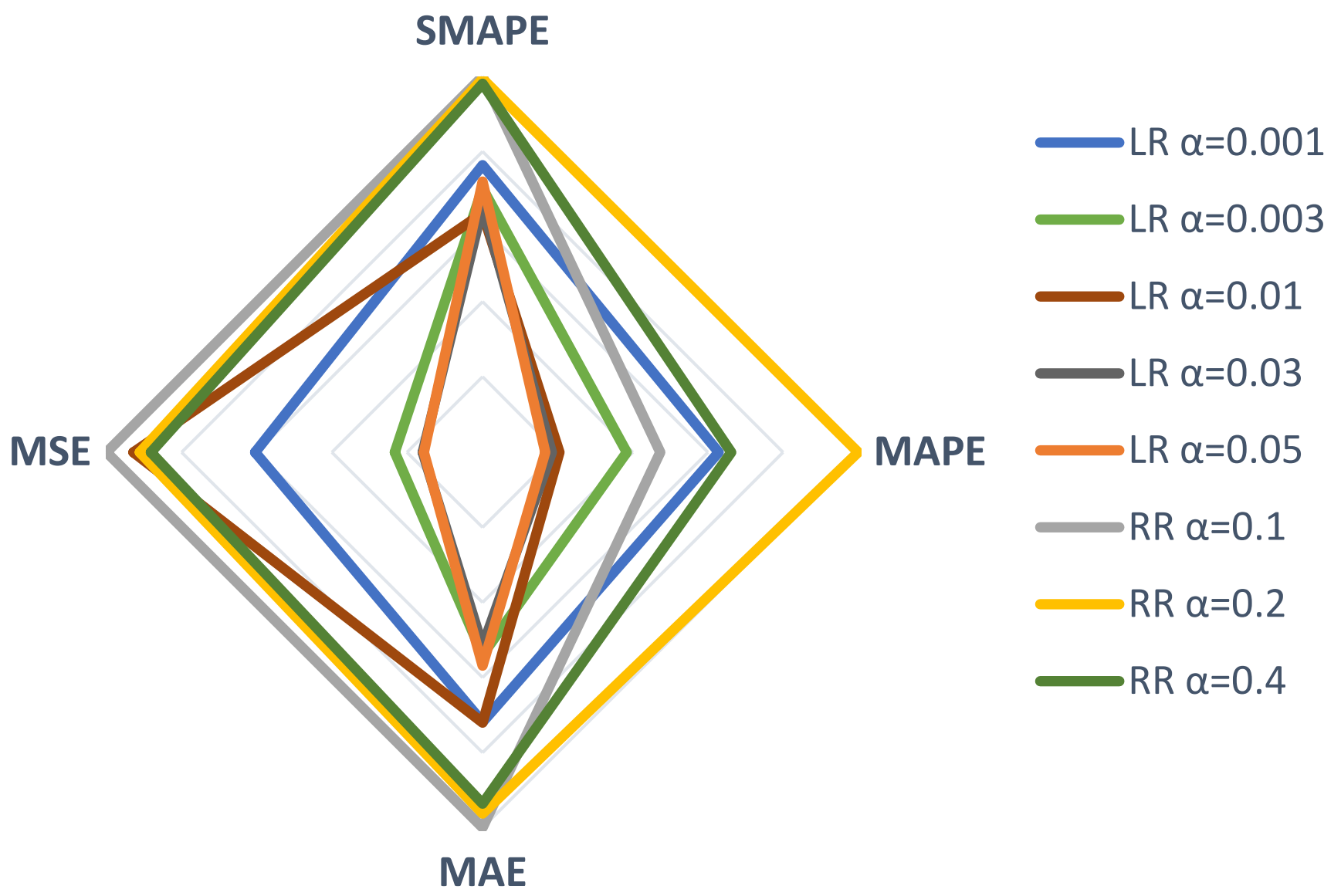

**Figure 26 – Radar plot comparing qualitatively the evaluation metrics for the selected estimators. Case study of a single well in a real oilfield.**

It is often more important characterizing the nuances and trends of the curves than simply obtaining the best error metrics, hence the importance of graphically observing the forecasts. This became clear when we noticed that Lasso regression with *α*=0.03 was more capable of generalizing than other configurations apparently superior in the measurements.

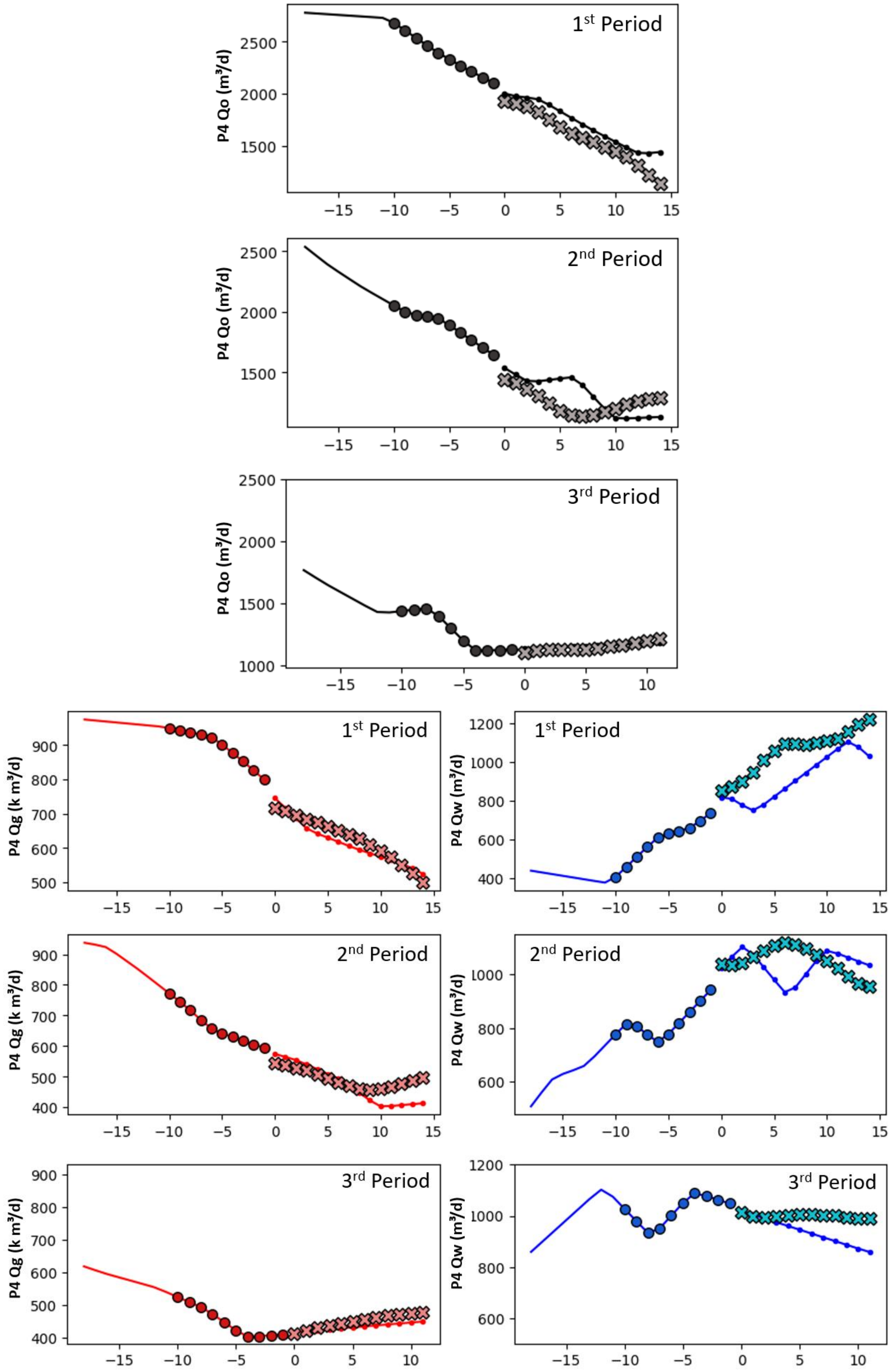

**Figure 27 – Oil, Gas and Water production rates for well P4 in the real case scenario, for three different timestamps. Forecast performed by Lasso Regression for 240 days (marked as X), using as inputs the previous 200 days (marked as circles).**

We present in Figure 27 the results for this best configuration, observing at three different moments and forecasting the next 240 days (more than our goal of 6 months). We note that the model can capture the general trends and some of the most subtle, but there is still difficulty in accurately reproducing the reference curves in such challenging scenario.

In real oilfields those challenges come partially from operational aspects, inaccuracy in measurements and volume appropriations, but mainly from the reservoir itself, including heterogeneities not represented in the models, creation of preferential paths in the porous medium after injection cycles, compositional variations of fluids, miscibility in relation to injected gases (with variable composition, especially in $CO_2$ content) and skin effects, in a non-exhaustive list. Difficulties that are meant to be represented in simulation models, but that are generally much more complex in real-world situations.

## 1.4 Conclusions

This work presents the development of a data-driven workflow for production forecasts of oilfields. The workflow includes important steps of data conditioning and the evaluation of machine learning methods, namely linear regression and neural networks. The results obtained are encouraging for the design of a reliable predictor, capable of providing fast responses, and based only on past production data and programmed injection rates, without the need to use information related to geology, rock/fluid interactions or even to wells and other production facilities.

We emphasize that we aim to forecast reliably for a minimum period of 6 months, what we consider adequate to practical applications as operational planning and short- and medium-term reservoir management decisions. The results of the case studies are in general satisfactory, and the models managed to reproduce general production tendencies even in challenging scenarios. Whan dealing with longer forecasts and transient behaviors, however, there is a margin for improvement.

Other relevant aspects are the great sensitivity to configurations and hyperparameters of learning methods and the need to obtain representative and reliable databases, given that these are the fundamental basis for any data-driven method.